\documentclass{ieeeaccess}

\usepackage{cite}
\usepackage{amsmath,amssymb,amsfonts}
\usepackage{bm}
\usepackage{graphicx}
\usepackage{algorithm}
\usepackage{algorithmic}

\vol{14}
\year{2026}
\begin{document}
\widowpenalty=10000

% Reserve the template's bullet and gap without changing the title width.
\newlength{\frontmattertextwidth}
\setlength{\frontmattertextwidth}{\dimexpr\titlewidth-0.5em-1.5mm\relax}
\let\abstractwidth\frontmattertextwidth

\history{This work has been submitted to the IEEE for possible publication. Copyright may be transferred without notice, after which this version may no longer be accessible.}
\doi{}
\title{Sector-Ring Linear Reservoirs: Sector-Wise Decay-Timescale Allocation With State Reuse Across Candidate Settings}
\author{\uppercase{Yudai Ebato}\authorrefmark{1},
\uppercase{Sou Nobukawa}\authorrefmark{2,3,4},
\uppercase{Yusuke Sakemi}\authorrefmark{3,5},
\uppercase{Takashi Kanamaru}\authorrefmark{6},\\
\uppercase{Nina Sviridova}\authorrefmark{5,7},
\uppercase{and Kazuyuki Aihara}\authorrefmark{3,5}}

\address[1]{Graduate School of Information and Computer Science, Chiba Institute of Technology, Narashino, Chiba 275-0016, Japan}
\address[2]{Department of Computer Science, Chiba Institute of Technology, Narashino, Chiba 275-0016, Japan}
\address[3]{Research Center for Mathematical Engineering, Chiba Institute of Technology, Narashino, Chiba 275-0016, Japan}
\address[4]{Department of Preventive Intervention for Psychiatric Disorders, National Institute of Mental Health, National Center of Neurology and Psychiatry, Kodaira, Tokyo 187-8551, Japan}
\address[5]{International Research Center for Neurointelligence (WPI-IRCN), The University of Tokyo, Bunkyo-ku, Tokyo 113-0033, Japan}
\address[6]{Department of Mechanical Science and Engineering, School of Advanced Engineering, Kogakuin University, Hachioji, Tokyo 192-0015, Japan}
\address[7]{Department of Intelligent Systems, Tokyo City University, Setagaya-ku, Tokyo 158-8557, Japan}
\tfootnote{This work was supported in part by a project, JPNP14004, commissioned by the New Energy and Industrial Technology Development Organization (NEDO); in part by JSPS KAKENHI Grant-in-Aid for Scientific Research (B) (Grant Numbers JP25K03198, SN; JP25K00148, YS); in part by Grant-in-Aid for Transformative Research Areas (A) (Grant Number JP25H02626, SN); and in part by Grant-in-Aid for JSPS Fellows (Grant Number JP25KJ2052, YE).}

\markboth
{Ebato \headeretal: Sector-Ring Linear Reservoirs}
{Ebato \headeretal: Sector-Ring Linear Reservoirs}

\corresp{Corresponding author: Yudai Ebato (e-mail: yudaiebato@gmail.com).}

\begin{abstract}
In reservoir computing, performance depends on whether the fixed dynamics of the reservoir suit the task, and the appropriate memory length and frequency response differ across tasks. The reservoir settings are found by a search on each task, and driving the reservoir with the input and collecting its states for every candidate setting dominates the cost of the search. In a reservoir with a linear recurrence, the dynamics decompose into modes, one per eigenvalue, and the eigenvalue distribution becomes a design variable. We propose the Sector-Ring Linear Reservoir (SRLR), in which the eigenvalue angles are fixed at evenly spaced positions around a circle and one decay timescale, chosen from a few values, is assigned to each sector of consecutive angles. To keep the scale of each mode's state equal across candidates, the input weights are scaled using the power spectrum of the training input. To reduce the cost of the search, the states of the modes for each decay-timescale value are computed once and reused across candidates. On prediction and reconstruction benchmarks built from chaotic systems, the SRLR achieved, in all eight cases, a lower mean test error than a reference model in which all modes share one decay timescale, and was competitive with nonlinear echo state network and leaky-integrator echo state network baselines. The selected decay-timescale profiles changed with the prediction target and horizon. The state reuse reduced the number of computed mode trajectories in the search by 99.8\% relative to per-candidate recomputation.
\end{abstract}


\titlepgskip=-21pt

\maketitle
\section{Introduction}
\label{sec:introduction}

Reservoir computing (RC) is a framework for time-series processing that uses a fixed, typically nonlinear, recurrent dynamical system called the reservoir~\cite{jaeger2001echo,maass2002real,lukosevicius2009reservoir}. The reservoir expands the input history into a high-dimensional state trajectory, and only a readout from that trajectory is trained, usually by linear regression. Echo State Networks (ESNs) and Liquid State Machines are representative examples~\cite{jaeger2001echo,maass2002real}. The framework has been applied effectively to time-series prediction, signal processing, and the prediction and reconstruction of dynamical systems, including chaotic ones~\cite{jaeger2004harnessing,pathak2018model}. Training the readout is a linear least-squares problem and is faster and cheaper than gradient-based training of recurrent networks, and this training efficiency is one of the main advantages cited for RC~\cite{lukosevicius2009reservoir,tanaka2019physical,lukosevicius2012practical}. On the other hand, because the recurrent dynamics are fixed, all the information the readout can use must already be present in the reservoir state. Performance therefore depends on whether the dynamics of the reservoir suit the task.

Useful reservoir dynamics are task-dependent. Two of the main factors that affect performance are the timescales over which the reservoir retains past input (its memory) and the frequency components of the input to which it responds (its frequency response). The amount of memory required differs with the prediction target~\cite{carroll2022optimizing}, and the timescales and frequency components to which the reservoir should respond have likewise been adjusted for each task~\cite{jaeger2007leaky,wyffels2008bandpass,tanaka2022diversetimescales}.

The hyperparameter settings that realize such dynamics are found by search on the task, and this search accounts for most of the computational cost of applying RC. Practical guides treat the main hyperparameters (the spectral radius, the leaking rate, and the input scaling) as quantities to be tuned together for the task~\cite{lukosevicius2012practical,jaeger2007leaky}, and applying RC to a task means evaluating many candidate settings on that task. In training an ESN, driving the reservoir with the input and collecting its states is the stage that dominates the cost, and the search repeats this stage for every candidate~\cite{lukosevicius2023crossvalidation}. The reservoir must be driven anew for each candidate because the recurrence is nonlinear and the effect of a setting appears only once the reservoir has been driven. This motivates a design variable that relates more directly to the recurrent dynamics and makes each candidate cheaper to evaluate.

One way to address this problem is to make the recurrence linear, that is, to use a linear reservoir. Linear systems theory then applies to the recurrence, and both the interpretation of the dynamics and the computation of the states become simpler. The state trajectory of a linear recurrence can be analyzed as a superposition of modes, components that evolve independently of one another, each corresponding to an eigenvalue~\cite{oppenheim1999discrete,nathe2021modal}. The radius (magnitude) of a stable discrete-time eigenvalue determines the decay timescale of its mode, that is, how long the mode retains past input, and the angle determines the oscillatory, frequency-like component of its response~\cite{oppenheim1999discrete,fong2025linearSCR}. On the computational side, the recurrences of the individual modes can be advanced in parallel across modes or by parallel scan algorithms over time~\cite{blelloch1990prefix,martin2018parallelizing}, as recent diagonal and parallel formulations of linear reservoirs demonstrate~\cite{decoudenhove2026linear,pinna2026paralesn}. The eigenvalue distribution thus becomes an explicit design variable that acts on each mode separately. Using this variable to search and compare candidate settings, however, requires further conditions.

Making the eigenvalue distribution a practical design variable for a reservoir requires three conditions. The first is that the eigenvalue distribution can be specified by a few parameters. If each eigenvalue is chosen freely~\cite{nathe2021modal,tangerami2025optimizing}, the search space is too large to search at low computational cost. The second is control of the scale, that is, the variance, of the state of each mode. Even under the same input, different eigenvalue settings give the state of each mode a different scale, and unless this variation is controlled, the differences among candidates include a factor other than the eigenvalue distribution. The third is that each candidate can be evaluated at low computational cost. The eigenvalue distribution adds parameters to tune on top of the existing hyperparameters; consequently, the number of candidates, and with it the number of state trajectories to compute, grows sharply. What is needed, therefore, is a design that specifies the eigenvalue distribution with a few parameters, keeps the scale of each mode's state equal across candidates, and can be searched at low computational cost.

In this paper, we propose the Sector-Ring Linear Reservoir (SRLR), a design that meets these conditions. The SRLR is a reservoir whose recurrent update is linear and in which nonlinearity is introduced by a fixed feature map before the readout. The angles of the modes' eigenvalues are fixed at evenly spaced positions around a circle, and one decay timescale is assigned to each sector of consecutive angles; a candidate is therefore specified by one value per sector, that is, by a few parameters. The input weight of each mode is scaled using the power spectrum of the training input so that the scale of each mode's state is equal across candidates. Moreover, because each per-sector value is chosen from a small set and the trajectory of each mode depends only on the value assigned to its sector, the trajectory of each mode can be computed once per value and reused across candidates. Our preliminary conference paper studied eigenvalue control in a linear reservoir (the eigenvalue-controlled ESN) using random eigenvalue distributions constrained to angular sectors~\cite{ebato2025ecesn}. The present paper develops that approach into the formulation just described: fixed angles, one decay timescale shared within each sector, scaling of the input weights using the power spectrum, and reuse of each mode's trajectory across candidates for each timescale value.

We evaluate the SRLR in two comparisons on prediction and reconstruction benchmark tasks built from chaotic dynamical systems. First, we test whether allowing the decay timescale to differ across sectors improves accuracy, by comparing the SRLR with the case in which all modes share one decay timescale, that is, the SRLR with a single sector. We call this case the common-radius reference. Second, we compare the SRLR with the ESN and leaky-integrator ESN (LI-ESN) baselines. We also report the selected decay-timescale profiles (the per-sector values chosen by the search), the scale of each mode's state, and the computational cost of the candidate search, and we compare the input-weight scaling with two alternative input-weight designs. The remainder of the paper is organized as follows. Section II reviews related work, Section III presents the models, tasks, and experimental protocol, Section IV reports the two comparisons and the additional analyses, Section V interprets the selected profiles and discusses the scope of the evidence, and Section VI concludes the paper.

\section{Related Work}
\label{sec:related-work}

\subsection{Task-Dependent Reservoir Dynamics}
\label{sec:related-task-dependent}

The amount of memory that a task requires is specific to that task. How long the memory of a reservoir lasts affects how well it solves a particular problem, and too much or too little memory can degrade accuracy~\cite{carroll2022optimizing}. Even among tasks that use the same input signal, the required memory changes with the prediction target and with the prediction horizon, that is, the number of steps ahead at which the target is predicted~\cite{jaurigue2024tailoring}. Moreover, the total ability of a reservoir to retain past inputs and to transform them nonlinearly, its information-processing capacity, is finite, and because memory and nonlinear transformation share it, memory cannot be increased without reducing nonlinear transformation~{\renewcommand{\citedash}{]\mbox{--}[}\cite{dambre2012information,inubushi2017beyond,sakemi2020modelsizereduction}}.

The timescales and the frequency response of a reservoir have also been adjusted to the task. Leaky integration sets a single timescale of the state update for the whole reservoir, and band-pass filters or filter neurons set the frequency response of each unit~\cite{jaeger2007leaky,wyffels2008bandpass,holzmann2010filterneurons}. There are also designs that give different timescales to individual neurons, to groups of neurons, or to sub-reservoirs~\cite{tanaka2022diversetimescales,yoshida2025heterogeneous,manneschi2021exploiting}, designs that set the decay timescale of each mode independently~\cite{beuria2026lindblad}, and designs that split the input into bands fed to separate reservoirs~\cite{li2023hpmresn}.

The dynamics of a reservoir are determined not only by its settings but also by the driving input. Even the echo state property, that is, the property that a driven reservoir forgets its initial condition, depends on the driving input~\cite{yildiz2012revisiting,manjunath2013echo}. Measures computed with generic test inputs, such as white noise, do not necessarily describe performance under the task input; indeed, the total information-processing capacity correlates only weakly with task performance~\cite{hulser2023deriving}. Moreover, in nonlinear reservoirs the measured memory capacity, which quantifies how well past inputs can be reconstructed from the state, changes with the input amplitude alone~\cite{jaeger2002shortterm,ballarin2025informative}. In view of this input dependence, memory has also been measured with the actual input signal; for example, the delay capacity quantifies memory using the autocorrelation of the states of the reservoir driven with the actual input~\cite{carroll2022optimizing,ebato2024impact}. Selecting the settings that suit a task therefore requires driving the reservoir with the input of that task and evaluating it for each candidate setting.

\subsection{Linear Reservoirs}
\label{sec:related-linear}

Cycle and ring reservoirs specify the eigenvalue distribution and the decay timescale with a few parameters, namely evenly spaced angles and one shared spectral radius, and such reservoirs have been shown to achieve accuracy comparable to that of randomly generated ESNs. The Simple Cycle Reservoir (SCR) couples nonlinear units in a ring through one shared weight, and because it uses no random weights it can be constructed deterministically. On benchmarks it achieved accuracy comparable to that of randomly generated ESNs~\cite{rodan2011minimum}. The single shared weight gives every eigenvalue the same magnitude, equal to the spectral radius; in the linear SCR all modes therefore share one decay timescale~\cite{tino2020dynamical,fong2025linearSCR}. The SRLR retains these evenly spaced angles and the small number of parameters, and its common-radius reference, like a linear SCR, has all eigenvalues at one magnitude, the spectral radius. The SRLR differs from these ring designs and from its own common-radius reference in that it varies the decay timescale across sectors.

A reservoir whose recurrence is linear and diagonalizable can be formulated in diagonal form, in which the recurrent matrix is diagonal and the eigenvalues of the individual modes are parameterized directly. In the diagonal form each mode evolves independently, one per eigenvalue, and the state trajectory is their superposition. This form has been used to analyze reservoirs by decomposition into modes~\cite{nathe2021modal} and, recently, as the architecture of linear reservoirs and of reservoirs whose diagonal recurrence has complex-valued entries (complex-diagonal ESNs)~\cite{decoudenhove2026linear,pinna2026paralesn}. In the complex-diagonal formulations the modes are parameterized independently of one another, and the decay timescale and the angle of each mode can be chosen separately.

A linear recurrence also has the advantage that its computation over time can be parallelized. Because each update in a linear recurrence is an affine transformation, a matrix product plus an input term, consecutive updates can be combined into one affine transformation and computed in parallel over time. A parallel scan algorithm called an associative scan uses this composition, and the number of sequential steps then scales logarithmically with the sequence length $T$~\cite{blelloch1990prefix,martin2018parallelizing}. In the diagonal form the matrix products reduce to element-wise products, which makes the composition computationally inexpensive, and this parallelization has yielded speedups for linear recurrent networks and has been applied to complex-diagonal ESNs~\cite{martin2018parallelizing,pinna2026paralesn}.

The combination of a linear reservoir with a nonlinear readout, that is, a fixed nonlinear map applied to the state before the trained linear readout, has been adopted in several lines of research and is supported both theoretically and empirically. Theoretically, linear recurrent dynamics followed by sufficiently expressive nonlinear maps are universal approximators for broad classes of sequence-to-sequence maps~{\renewcommand{\citedash}{]\mbox{--}[}\cite{gonon2020stochastic,orvieto2024universality,li2024simple,fong2025real}}. Empirically, competitive accuracy in this form has been reported for next-generation reservoir computing, which reads out nonlinear functions of time-delayed inputs~\cite{gauthier2021next}, for nonlinear readouts in linear reservoir networks in a benchmark study~\cite{lagomarsini2026benchmarking}, and for complex-diagonal ESNs~\cite{pinna2026paralesn}.

The choice of eigenvalues also changes the scale of each mode's state. Per-mode normalizations have therefore been proposed that scale the input weight of each mode so that the variance of each mode's state is equal across modes, as in the normalization parameter $\gamma$ of the Linear Recurrent Unit (LRU)~\cite{orvieto2023resurrecting} and the per-mode scaling of complex-diagonal ESNs~\cite{pinna2026paralesn}. Because these assume a white-noise input, we scale the input weights using the power spectrum of the training input so that different eigenvalue distributions can be compared with the state variance of each mode held equal across candidates.

\section{Methods}

This section describes the reservoir models and the common framework in which
they are evaluated (Section~III-A), and the benchmark tasks, the evaluation
protocol, and the candidate-evaluation computation of the SRLR (Section~III-B). The training, validation, and test windows and the model seeds referred to in Section~III-A are defined by the evaluation protocol of Section~III-B2; a model seed is the random seed that fixes the random elements of a model, namely the weights of the ESN and the LI-ESN (Section~III-A2), the mixing matrix of the SRLR (Section~III-A3d), and the input weights of the uniform random input-weight design of the SRLR (Section~III-A3c).

We conduct two comparisons. The first tests whether allowing the decay
timescale to differ across sectors improves accuracy, by comparing the
SRLR with its common-radius reference. The second tests how the
SRLR as a whole, with its linear recurrence, compares with the nonlinear
ESN and LI-ESN baselines. In addition, we compare the input-weight scaling of
the SRLR with two alternative input-weight designs (Section~III-A3c),
each under its own independent search.

\subsection{Reservoir-Model Design and Evaluation Framework}

\subsubsection{Common Reservoir-Evaluation Framework}

This subsection describes the common framework in which all reservoir models
are compared. The models compared are the ESN, the LI-ESN, the Sector-Ring Linear Reservoir (SRLR), and the common-radius reference, which is the SRLR with a single sector (Section~III-A3b). Each
model updates its reservoir state \(\mathbf h_t \in \mathbb R^N\) from the
preprocessed input \(\widetilde u_t\) (Section~III-B2) by its own recurrence
and forms a feature vector \(\boldsymbol\chi_t \in \mathbb R^{N+1}\) by a fixed feature map, where \(N\) is the number of reservoir units. In every model the feature vector \(\boldsymbol\chi_t\) consists of \(N\) entries formed from the state, followed by a constant entry of \(1\) (the bias entry); its specific form is given in the subsection for each model. The feature vector is mapped to predictions by a readout
\(\hat{\mathbf y}_t = \mathbf W_{\mathrm{out}}^\mathsf T \boldsymbol\chi_t\),
trained for each model by the same ridge regression, and prediction quality is
scored by the normalized root-mean-square error (NRMSE) under a protocol
shared by all models (Section~III-B).

Only the readout weights are trained, by ridge regression. For each model, the feature vectors over the training window (of length \(T_{\mathrm{tr}}\) samples, Section~III-B2) are stacked as rows into the feature matrix \(\boldsymbol\Phi_{\mathrm{tr}} \in \mathbb R^{T_{\mathrm{tr}} \times (N+1)}\),
and the corresponding targets into
\(\mathbf Y_{\mathrm{tr}} \in \mathbb R^{T_{\mathrm{tr}} \times d_y}\), where
\(d_y\) is the target dimension (\(d_y = 1\) for all tasks in this paper,
Section~III-B1). The readout weights
\(\mathbf W_{\mathrm{out}} \in \mathbb R^{(N+1) \times d_y}\) are then
obtained in closed form as
\begin{equation}
\mathbf W_{\mathrm{out}}
= (\boldsymbol\Phi_{\mathrm{tr}}^\mathsf T \boldsymbol\Phi_{\mathrm{tr}}
+ \beta \mathbf I)^{-1}
\boldsymbol\Phi_{\mathrm{tr}}^\mathsf T \mathbf Y_{\mathrm{tr}},
\label{eq:ridge}
\end{equation}
where \(\beta > 0\) is the ridge regularization parameter and \(\mathbf I\) is
the identity matrix. Because the last entry of \(\boldsymbol\chi_t\) is the constant \(1\), the last column of \(\boldsymbol\Phi_{\mathrm{tr}}\) is a column of ones, and the last row of \(\mathbf W_{\mathrm{out}}\) is the bias of the readout; this bias is regularized by the same \(\beta\) in (\ref{eq:ridge}), and no separate unregularized intercept is fitted. The parameter \(\beta\) is one of the model-specific settings selected on the validation window (Section~III-B2).

\subsubsection{ESN and LI-ESN Baselines}

The first baseline model is the ESN~\cite{jaeger2001echo}. The ESN updates the state
\(\mathbf h_t \in \mathbb R^N\) of its \(N\) reservoir units from the
preprocessed scalar input \(\widetilde u_t\) as
\begin{equation}
\mathbf h_t = \tanh(\mathbf W_{\mathrm{rec}} \mathbf h_{t-1}
+ \mathbf w_{\mathrm{in}} \widetilde u_t),
\qquad \mathbf h_0 = \mathbf 0,
\label{eq:esn}
\end{equation}
where \(\tanh\) is applied element-wise,
\(\mathbf w_{\mathrm{in}} \in \mathbb R^N\) is the input weight vector, and
\(\mathbf W_{\mathrm{rec}} \in \mathbb R^{N \times N}\) is the recurrent
weight matrix. For each model seed (Section~III-B2), each input weight \(w_{\mathrm{in},i}\) is drawn uniformly from \((-s_{\mathrm{in}}, s_{\mathrm{in}})\), where \(s_{\mathrm{in}}\) is the input scaling. The recurrent weight matrix
\(\mathbf W_{\mathrm{rec}}\) is a sparse random matrix constructed as follows.
A matrix \(\widetilde{\mathbf W}\) with entries
\(\widetilde W_{ij} = \xi_{ij} Z_{ij}\) is first formed by drawing
\(\xi_{ij}\) from \(\mathrm{Bernoulli}(0.1)\) and \(Z_{ij}\) uniformly from
\((-1, 1)\). The matrix \(\widetilde{\mathbf W}\) is then divided by its spectral radius
\(\rho(\widetilde{\mathbf W})\), the largest absolute eigenvalue, and
multiplied by \(r\), giving
\(\mathbf W_{\mathrm{rec}} = (r/\rho(\widetilde{\mathbf W}))\,\widetilde{\mathbf W}\),
whose spectral radius is therefore \(r\). For each model seed, a single random stream draws the input weights first and then the entries \(\xi_{ij}\) and \(Z_{ij}\) of \(\widetilde{\mathbf W}\); no per-row rescaling of the recurrent weights is applied, so that the spectral-radius scaling above is the only scaling of \(\mathbf W_{\mathrm{rec}}\). The state update (\ref{eq:esn}) has no bias term and no output feedback~\cite{lukosevicius2012practical}, that is, the readout output is not fed back into the state. The feature vector is the state with the bias entry appended, \(\boldsymbol\chi_t = [\mathbf h_t^\mathsf T, 1]^\mathsf T \in \mathbb R^{N+1}\). The hyperparameters selected on the validation window are the input scaling
\(s_{\mathrm{in}}\), the spectral radius \(r\), and the ridge parameter
\(\beta\).

The second baseline model is the LI-ESN~\cite{jaeger2007leaky,lukosevicius2012practical}. It differs from the ESN in the state update,
\begin{equation}
\begin{gathered}
\mathbf h_t = (1 - \alpha_\ell)\,\mathbf h_{t-1}
+ \alpha_\ell \tanh(\mathbf W_{\mathrm{rec}} \mathbf h_{t-1}
+ \mathbf w_{\mathrm{in}} \widetilde u_t),\\
0 < \alpha_\ell \le 1,
\end{gathered}
\label{eq:liesn}
\end{equation}
where the leaking rate \(\alpha_\ell\) adjusts the timescale of the state update: the new state is a weighted average of the previous state and the tanh update. The LI-ESN uses the same weight construction and feature vector as the ESN (the spectral radius \(r\) is that of \(\mathbf W_{\mathrm{rec}}\), as in the ESN) and adds the leaking rate \(\alpha_\ell\) to the hyperparameters selected on the validation window.

\subsubsection{Sector-Ring Linear Reservoir}

\begin{figure*}[t]
\centering
\includegraphics[width=\textwidth]{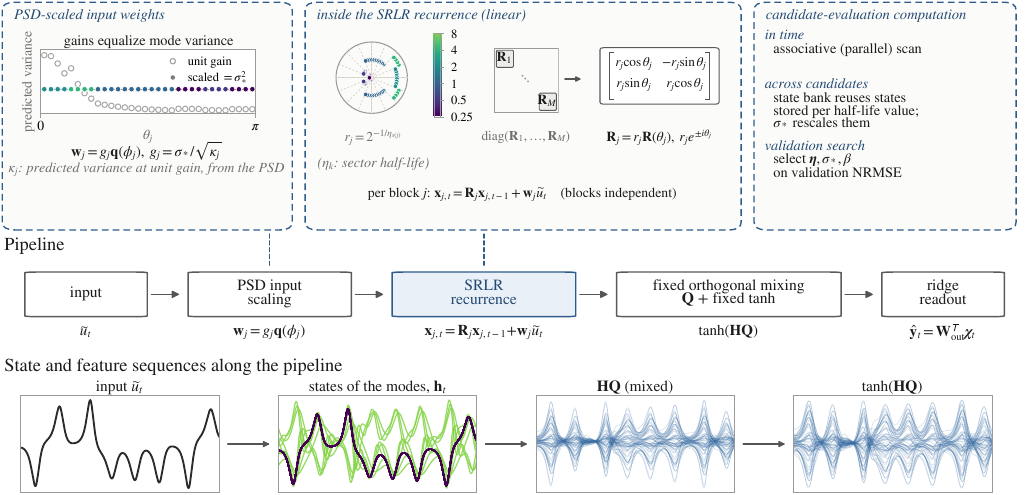}
\caption{Concept of the Sector-Ring Linear Reservoir. The preprocessed input
\(\widetilde u_t\) enters through input weights scaled using the power spectral density (PSD)
of the training input (Section~III-A3c). Rotation--decay blocks assigned to
sectors generate the states of the modes, which a fixed orthogonal mixing and a
fixed tanh map transform into features for the ridge readout. Here
\(\mathbf q(\phi) = [\cos\phi, \sin\phi]^\mathsf T\) denotes the unit
direction of the input weights of one mode.
In the pipeline and the bottom row, \(\mathbf H\) denotes the states \(\mathbf h_t\) stacked
in rows over time, so that the rows of \(\mathbf H \mathbf Q\) are the mixed
states \(\mathbf Q^\mathsf T \mathbf h_t\) (Section~III-B3).
The upper-right callout summarizes the candidate-evaluation computation (Section~III-B3): a state bank indexed by the
half-life value (the per-sector decay timescale, Section~III-A3b) reuses
states across candidates. An associative scan is invoked only for a half-life
value that the candidate requires and the bank does not yet hold.}
\label{fig:method-sector-ring-concept}
\end{figure*}

\paragraph{Rotation--Decay Block Representation}

As shown in Fig.~\ref{fig:method-sector-ring-concept}, the SRLR
implements its recurrent matrix as a block-diagonal matrix of \(2 \times 2\)
rotation--decay blocks, so that the eigenvalue distribution of a real
recurrent matrix can be designed directly. With \(N = 2M\), block \(j\)
(\(j = 1, \ldots, M\)) holds a state \(\mathbf x_{j,t} \in \mathbb R^2\)
driven by the preprocessed scalar input \(\widetilde u_t\) through the input
weights \(\mathbf w_j \in \mathbb R^2\), and is updated as
\begin{equation}
\begin{gathered}
\mathbf x_{j,t} = \mathbf R_j \mathbf x_{j,t-1} + \mathbf w_j \widetilde u_t, \qquad \mathbf x_{j,0} = \mathbf 0,\\
\mathbf R_j = r_j \mathbf R(\theta_j),
\quad
\mathbf R(\theta) =
\begin{bmatrix}
\cos\theta & -\sin\theta \\
\sin\theta & \cos\theta
\end{bmatrix},
\end{gathered}
\label{eq:block}
\end{equation}
where \(\mathbf R(\theta)\) is the \(2 \times 2\) rotation matrix. Each block
realizes one complex-conjugate eigenvalue pair \(r_j e^{\pm i\theta_j}\),
where \(r_j\) is the amplitude decay factor per update and \(\theta_j\) the
phase increment per update. Below, the state \(\mathbf x_{j,t}\) of block
\(j\) is called the state of mode \(j\) (a mode in the sense of Section~I, a
component that evolves independently, and in this real form one mode carries
one conjugate eigenvalue pair). Because the decay of each mode is specified
directly by the radius \(r_j\), leaky integration as in the LI-ESN is not used.
The reservoir state is the concatenation of the states of the modes,
\(\mathbf h_t = [\mathbf x_{1,t}^\mathsf T, \ldots,
\mathbf x_{M,t}^\mathsf T]^\mathsf T \in \mathbb R^N\).

\paragraph{SRLR Eigenvalue Parameterization and Common-Radius Reference}

We now describe the eigenvalue distribution of the SRLR. The angles are
fixed at the midpoints
\begin{equation}
\theta_j = \Bigl(j - \tfrac{1}{2}\Bigr)\frac{\pi}{M},
\qquad j = 1, \ldots, M,
\label{eq:angles}
\end{equation}
of the \(M\) equal subintervals of \((0, \pi)\). The conjugate eigenvalue of
each pair lies at angle \(-\theta_j\), and the eigenvalue pairs are therefore
symmetric about the real axis. Placing the angles at the midpoints avoids the
angles \(0\) and \(\pi\), at which \(\mathbf R(\theta) = \pm\mathbf I\) and the block would have a repeated real eigenvalue, so that its two coordinates would carry the same response up to a scale factor. The interval
\((0, \pi)\) is then divided into \(K\) consecutive sectors of equal width, and
block \(j\) belongs to sector
\begin{equation}
s(j) = \left\lfloor \frac{K\theta_j}{\pi} \right\rfloor
\in \{0, \ldots, K-1\},
\label{eq:sector}
\end{equation}
where \(\lfloor\cdot\rfloor\) is the floor function
(Fig.~\ref{fig:method-sector-ring-concept}). The angles and the sector membership are thus fixed by (\ref{eq:angles}) and (\ref{eq:sector}), with no random draw. Below, \(k\) denotes a sector
index, and \(s(j)\) denotes the index of the sector containing block \(j\).
When \(M\) is a multiple of \(K\), each sector contains exactly \(M/K\) modes.
In the experiments of this paper, \(N = 64\), \(M = 32\), and \(K = 8\), so that each
sector contains four modes. The \(k\)-th sector is assigned one positive
half-life \(\eta_k\), and the vector
\(\boldsymbol\eta = (\eta_0, \ldots, \eta_{K-1})\) of the sector half-lives is
called the half-life profile (the decay-timescale profile of Section~I). The
half-life is the number of updates over which the amplitude of a mode decays
by half, and it serves as the decay timescale of the mode. In terms of a time
constant, the amplitude decays as \(e^{-t/\tau_k}\) with
\(\tau_k = \eta_k / \ln 2\). The radius of block \(j\) is determined by the
half-life of its sector as
\begin{equation}
r_j = 2^{-1/\eta_{s(j)}},
\label{eq:radius}
\end{equation}
so that its amplitude halves every \(\eta_{s(j)}\) updates. Since
\(\eta_{s(j)} > 0\) gives \(0 < r_j < 1\), every eigenvalue lies inside the
unit circle. When the reservoir is driven with the same input from any two
initial states, the difference between the two state trajectories vanishes
asymptotically, which gives this linear recurrence the echo state property~\cite{manjunath2013echo}.

The common-radius reference is the special case of the SRLR with a
single sector, \(K = 1\), in which all blocks share one half-life \(\eta_0\)
and hence one radius \(r = 2^{-1/\eta_0}\). As in a linear SCR, all of its
eigenvalues have the same magnitude \(r\) and evenly spaced angles
(Section~II-B). The rest of the construction, namely the fixed angles, the
input-weight scaling, the fixed feature map, and the common ridge readout and
scoring, is unchanged. The search spaces and computational budgets of the two
models are described in Section~III-B2.

\paragraph{Input-Weight Scaling From the Training-Input Power Spectrum}

In a linear reservoir such as the SRLR, the input weights require attention
in addition to the eigenvalues. Mode \(j\) responds most strongly to input
components whose normalized angular frequency is close to \(\theta_j\)~\cite{oppenheim1999discrete}, and the variance of its state is determined by
the power spectral density (PSD) of the preprocessed training input weighted
by the frequency response of block \(j\). Consequently, the scale of each
mode's state changes with the eigenvalue setting. Existing normalizations that
assume white-noise input~\cite{orvieto2023resurrecting} (Section~II-B) do not
account for this frequency-selective response to an input that is not white.
We therefore scale the input weights using the PSD of the preprocessed
training input.

To control this variation in scale across candidate eigenvalue distributions,
the input weights are set analytically so that the states of all modes have a
common predicted scale. The input weights of block \(j\) are written as the
product of a scalar gain \(g_j\) and a unit direction \(\mathbf q(\phi_j)\),
\begin{equation}
\mathbf w_j = g_j \mathbf q(\phi_j),
\qquad
\mathbf q(\phi) =
\begin{bmatrix}
\cos\phi \\
\sin\phi
\end{bmatrix}.
\label{eq:input-weights}
\end{equation}
For each block, the direction \(\phi_j\) is determined from the PSD of the preprocessed training input so that the two coordinates of the mode's state
have equal predicted variance. With \(\kappa_j\) (\(> 0\)) denoting the
predicted variance at unit gain and \(\sigma_*\) the target scale (the target
standard deviation of each coordinate), the gain is
\(g_j = \sigma_*/\sqrt{\kappa_j}\), which gives both coordinates the target
variance \(\sigma_*^2\). In this prediction, the variance of the state of mode
\(j\) is obtained by integrating the one-sided Welch PSD~\cite{welch1967use} of the preprocessed
training input, weighted by the frequency response of block \(j\), over
\([0, f_s/2]\), where \(f_s\) is the sampling frequency of the input; in the derivation the frequency axis is normalized to \(f_s = 2\pi\), so that \(f\) is the normalized angular frequency in radians per update. The derivation is given in Appendix~\ref{app:input-weight-scaling}.

To distinguish the designs in the comparison, the design above is referred to as the PSD-scaled design, and
two alternative input-weight designs are also evaluated in the same
SRLR, with the fixed angles, the sector-wise half-life
parameterization, the fixed feature map, and the ridge readout unchanged. The
\(\gamma\)-normalized design, named after the \(\gamma\) normalization of~\cite{orvieto2023resurrecting}, determines the direction \(\phi_j\) and the
gain \(g_j\) by the same procedure as above but uses, instead of the PSD of
the training input, the flat one-sided PSD of unit-variance white noise,
\(P(f) = 2/f_s\). Its input weights therefore do not depend on the training
input, and they equalize the scale of each mode's state under the assumption
of white-noise input, the assumption made by the normalizations cited in
Section~II-B. The uniform random design draws each entry of the input weights
of all modes uniformly from \((-1, 1)\) once per model seed, holds them fixed,
and multiplies them by an input scaling \(s_{\mathrm{in}}\). Because this
design has no \(\sigma_*\) and no per-mode variance adjustment,
\(s_{\mathrm{in}}\) is selected on the validation window, as for the ESN
(Section~III-B2).

\paragraph{Fixed Nonlinear Feature Map and Model Settings}

Unlike the ESN and the LI-ESN, the SRLR has a linear recurrent state update, and its nonlinearity is introduced by a fixed feature map (Section~II-B). The feature map uses
\(\tanh\), the same function that the ESN and LI-ESN use inside their
recurrence. Here it is applied once, outside the recurrence, to a fixed orthogonal mixing of the state to form the feature vector, which is
\begin{equation}
\boldsymbol\chi_t = [\tanh(\mathbf Q^\mathsf T \mathbf h_t)^\mathsf T, 1]^\mathsf T
\in \mathbb R^{N+1},
\qquad \mathbf Q^\mathsf T \mathbf Q = \mathbf I,
\label{eq:features}
\end{equation}
where \(\mathbf Q \in \mathbb R^{N \times N}\) is a fixed orthogonal mixing
matrix, \(\tanh\) is applied element-wise, and the final entry \(1\) is the bias entry.

The fixed orthogonal matrix \(\mathbf Q\) supplies the mixing between modes, so that each feature is a nonlinear function of many modes rather than of a single mode.
Because the modes do not interact in the block-diagonal recurrence,
\(\mathbf Q^\mathsf T\) instead maps the reservoir state \(\mathbf h_t\) into
dense mixed coordinates, preserving its Euclidean norm, before the
element-wise \(\tanh\). The matrix \(\mathbf Q\) is generated once per model seed as a
Haar-distributed orthogonal matrix (the \(\mathbf Q\) factor of the QR
decomposition of a standard-normal matrix, with its column signs fixed), and
is then held fixed, together with the tanh map, across all candidate
evaluations of that seed.

Overall, three settings determine an SRLR model: the half-life profile
\(\boldsymbol\eta\) determines the sector radii of the recurrence, the target
scale \(\sigma_*\) determines the input-weight scaling, and \(\beta\)
determines the ridge readout. The angles, the feature map, and the scoring
procedure are held fixed. For the uniform random input-weight design
(Section~III-A3c), the input scaling \(s_{\mathrm{in}}\) replaces
\(\sigma_*\) among the three settings.

\subsection{Benchmark Design and SRLR Candidate Evaluation}

\subsubsection{Benchmark Time Series and Tasks}

We evaluate all models on supervised prediction and reconstruction tasks built
from the Lorenz~\cite{lorenz1963deterministic} and R\"ossler~\cite{rossler1976equation} systems. The Lorenz system is
\[
\begin{aligned}
\dot x_{\mathrm L} &= \sigma_{\mathrm L}(y_{\mathrm L} - x_{\mathrm L}), \\
\dot y_{\mathrm L} &= x_{\mathrm L}(\rho_{\mathrm L} - z_{\mathrm L})
                       - y_{\mathrm L}, \\
\dot z_{\mathrm L} &= x_{\mathrm L} y_{\mathrm L}
                       - \beta_{\mathrm L} z_{\mathrm L},
\end{aligned}
\]
with \(\sigma_{\mathrm L} = 10\), \(\rho_{\mathrm L} = 28\),
\(\beta_{\mathrm L} = 8/3\), and the R\"ossler system is
\[
\begin{aligned}
\dot x_{\mathrm R} &= -y_{\mathrm R} - z_{\mathrm R}, \\
\dot y_{\mathrm R} &= x_{\mathrm R} + a_{\mathrm R} y_{\mathrm R}, \\
\dot z_{\mathrm R} &= b_{\mathrm R} + z_{\mathrm R}(x_{\mathrm R} - c_{\mathrm R}),
\end{aligned}
\]
with \(a_{\mathrm R} = 0.15\), \(b_{\mathrm R} = 0.2\),
\(c_{\mathrm R} = 10\). Each system is integrated using a fixed-step
fourth-order Runge--Kutta method from the initial state
\((x_S(0), y_S(0), z_S(0)) = (1,1,1)\), \(S \in \{\mathrm L, \mathrm R\}\),
with an internal step of \(0.002\) and a sampling interval of \(0.02\). After discarding the initial point and an initial transient of 2,000 saved samples, one trajectory of 22,012 samples is retained. For each system \(S\) we build four observed-input tasks
that differ in the prediction target and the horizon:
\(x_S(t) \to x_S(t+1)\), \(x_S(t) \to x_S(t+12)\), \(x_S(t) \to z_S(t)\), and
\(x_S(t) \to z_S(t+12)\), giving eight cases in total. Here the horizon is the
number of saved samples from the input to the target, and horizons 1 and 12
correspond to \(0.02\) and \(0.24\) continuous-time units. Below, the \(x_S \to x_S\) tasks are called prediction tasks, and the \(x_S \to z_S\) tasks, whose target is a variable that the model does not observe, are called reconstruction tasks; the task \(x_S(t) \to z_S(t)\) is reconstruction at horizon 0. All tasks are
open-loop: the model always receives the observed input \(x_S(t)\), and its
predictions are never fed back.

\subsubsection{Training, Validation, Candidate Search, and Test Protocol}

The input time series of each system is divided chronologically into warmup,
training, validation, and test windows. In terms of the time index of the retained trajectory of Section~III-B1, these are
\([0, 2000)\) for reservoir warmup, \([2000, 12000)\) for training,
\([12000, 17000)\) for validation, and \([17000, 22000)\) for test, with the
final samples \([22000, 22012)\) retained only as a target margin for the
12-step horizon. All preprocessing statistics are computed from the training
window alone. The input is standardized by its training mean and standard
deviation, and this transform is applied to the warmup, training, and
validation and test inputs alike. The targets are standardized by their training
statistics, and this transform is applied to the training targets and to the
targets of the scored window. This defines the preprocessed input
\(\widetilde u_t\) used by all models. Every run resets the reservoir state to
zero and drives the reservoir with the 2,000 warmup inputs without scoring, so that the influence of the zero initial state has decayed before states are collected (the echo state property, Section~II-A).
States are then collected over the training window, and the run continues chronologically into one of the two scored passes (validation or test) described below. One
fixed trajectory and one fixed chronological split are used per system, with
no cross-validation.

The hyperparameters of each model are tuned separately for every task and model seed. All reservoir models use \(N = 64\) units, and the search coordinates (the hyperparameters tuned by the search) are \((s_{\mathrm{in}}, r, \beta)\) for the ESN and
\((s_{\mathrm{in}}, r, \alpha_\ell, \beta)\) for the LI-ESN. An SRLR
candidate is the tuple \(\mathbf c = (\sigma_*, \boldsymbol\eta, \beta)\),
where the half-life profile \(\boldsymbol\eta\) assigns one half-life to each
of the \(K\) sectors. For the two alternative input-weight designs of the
SRLR (Section~III-A3c), the candidate is
\((\sigma_*, \boldsymbol\eta, \beta)\) for the \(\gamma\)-normalized design
and \((s_{\mathrm{in}}, \boldsymbol\eta, \beta)\) for the uniform random
design, with the same range of \(s_{\mathrm{in}}\) as for the ESN. The
common-radius reference is searched in its own coordinates
\((\sigma_*, r, \beta)\). The shared radius \(r\) is one continuous parameter, and this model is the \(K = 1\) special case of Section~III-A3b. Its range of \(r\) in Table~\ref{tab:search-spaces}, \([2^{-4}, 2^{-1/8}]\), corresponds under (\ref{eq:radius}) to half-lives from 0.25 to 8, the smallest and largest values allowed for \(\eta_k\).
For every system, task, and model seed \(m = 0, \ldots, 9\), we run an
independent hyperparameter search using the Tree-structured Parzen Estimator
(TPE)~\cite{bergstra2011algorithms}, a sequential model-based optimizer that proposes each new candidate from the validation scores of the candidates evaluated so far, implemented in the hyperparameter-optimization library Optuna~\cite{akiba2019optuna}, with sampler seed \(42 + m\). The six search
configurations, namely the four models and the two alternative input-weight
designs of the SRLR (Section~III-A3c), use the same evaluation-budget
schedule and stopping rule. Each search starts with 2,048 evaluations (one evaluation scores one candidate on the validation window) and is extended in 1,024-evaluation increments until the stopping rule below is met
or the search reaches 8,192 evaluations. The searches remain independent
across systems, tasks, and model seeds, and stopping is decided for each search configuration and system--task case on the basis of all ten of its model seeds;
consequently, the ten searches of one search configuration and case end at the
same evaluation count, which may differ between cases. The search coordinates
and domains of each search configuration are summarized in
Table~\ref{tab:search-spaces}, and the realized evaluation counts are given
per case in Appendix~\ref{app:validation-search-progress}. The optimization
histories of these searches are shown in
Appendix~\ref{app:validation-search-progress}.

\begin{table*}[t]
\renewcommand{\arraystretch}{1.15}
\caption{Search spaces. Realized evaluation counts are given per case in Appendix~\ref{app:validation-search-progress}. Here (lin) and (log) denote uniform sampling on a linear and a logarithmic scale, respectively.}
\label{tab:search-spaces}
\centering
\begin{tabular}{p{0.28\textwidth} p{0.62\textwidth}}
\hline
Search configuration & Coordinates and domains \\
\hline
\rule{0pt}{2.8ex}ESN &
\(s_{\mathrm{in}} \in [0.1, 3.0]\) (lin),
\(r \in [0.2, 1.5]\) (lin),
\(\beta \in [10^{-9}, 10^{-3}]\) (log) \\
LI-ESN &
ESN coordinates plus \(\alpha_\ell \in [0.001, 1]\) (lin) \\
SRLR, PSD-scaled input weights &
\(\sigma_* \in [0.25, 4]\) (lin), \(\beta\) (log, as above), each
\(\eta_k \in \{0.25, 0.5, 1, 2, 4, 8\}\) \\
Common-radius reference &
\(\sigma_*\), \(\beta\) as above, shared
\(r \in [2^{-4}, 2^{-1/8}]\) (approximately \([0.0625, 0.9170]\); lin,
continuous) \\
SRLR, \(\gamma\)-normalized input weights &
\(\sigma_*\), \(\beta\), \(\boldsymbol\eta\) as above \\
SRLR, uniform random input weights &
\(s_{\mathrm{in}} \in [0.1, 3.0]\) (lin), \(\beta\), \(\boldsymbol\eta\) as
above \\
\hline
\multicolumn{2}{p{0.93\textwidth}}{\footnotesize
Note: The evaluated SRLR has \(K = 8\).}
\end{tabular}
\end{table*}

The stopping rule is as follows. For each search, let \(E^*(n)\) denote the
lowest validation NRMSE among the first \(n\) evaluations, and define the
relative improvement over the previous \(\Delta n = 1{,}024\) evaluations as
\begin{equation}
\delta(n) = \frac{E^*(n - \Delta n) - E^*(n)}{E^*(n - \Delta n)}.
\label{eq:stopping}
\end{equation}
The rule is checked at \(n = 3{,}072, 4{,}096, \ldots, 8{,}192\), and the
searches of a case stop at the first \(n\) at which the median of
\(\delta(n)\) over the ten model seeds of that case falls below \(0.5\%\). The
searches of a case that has not met the rule at 8,192 evaluations are stopped
at that count and reported as having reached the evaluation budget.

Model settings are selected on the validation window and then evaluated once
on the test window with all settings fixed. During the search, each candidate
is scored on the validation window: the state is reset, the reservoir is
driven with the warmup inputs without scoring, the readout is fitted on the
training window, and the run continues into the validation window. For each
case and model seed, the candidate with the lowest finite validation NRMSE is
selected (Section~III-B4). The final test repeats the same chronological pass
with the selected settings and no tuning. The reservoir is driven with the
validation inputs without scoring so that its state remains synchronized with
the input sequence, and the NRMSE is computed on the test window only.

\subsubsection{SRLR Candidate-Evaluation Computation}

Because the SRLR searches over eigenvalue distributions, the number of
candidate combinations is large and the search is computationally expensive.
We therefore reduce this cost by reusing states across candidates through a
state bank indexed by the discrete half-life values allowed for \(\eta_k\) in
Table~\ref{tab:search-spaces} and kept over one search (one system, task, and
model seed). For a candidate \(\mathbf c = (\sigma_*, \boldsymbol\eta, \beta)\)
of the PSD-scaled design, Algorithm~\ref{alg:candidate-evaluation} returns its
validation NRMSE using the state bank \(\mathcal B\) of the current search,
which starts empty and is kept across candidates, and the mixing matrix
\(\mathbf Q\) of the model seed. The states for a half-life value are
generated at \(\sigma_* = 1\) by an associative scan only when the bank does
not yet hold them (lines 1--7), and the candidate is then assembled from the
stored states (lines 8--11). In Algorithm~\ref{alg:candidate-evaluation},
\(\mathbf H_{\mathrm{tr}} \in \mathbb R^{T_{\mathrm{tr}} \times N}\) and
\(\mathbf H_{\mathrm v} \in \mathbb R^{T_{\mathrm v} \times N}\) are the
states \(\mathbf h_t\) stacked in rows over the training window and the
validation window (of length \(T_{\mathrm v}\) samples), so that the row form
\(\mathbf H \mathbf Q\) corresponds to \(\mathbf Q^\mathsf T \mathbf h_t\) in
(\ref{eq:features}). The matrix \(\mathbf Y_{\mathrm v} \in \mathbb R^{T_{\mathrm v} \times d_y}\) denotes the validation targets stacked in the same way, and \(\mathbf 1\) in
line 12 is a column of ones.

\begin{algorithm}[t]
\caption{SRLR candidate evaluation with a state bank.}
\label{alg:candidate-evaluation}
\begin{algorithmic}[1]
\FOR{each distinct half-life value \(\ell\) in \(\boldsymbol\eta\)}
  \IF{\(\mathcal B\) has no entry for \(\ell\)}
    \STATE Set the half-life of every sector to \(\ell\) and \(\sigma_*\) to
      \(1\), and compute the input weights \(\mathbf w_j\) of all \(M\) modes
      from the training-input PSD by (\ref{eq:input-weights})
      (Section~III-A3c).
    \STATE From the zero state, drive the reservoir with the warmup, training,
      and validation inputs in sequence by the associative scan
      (Section~III-B3a).
    \STATE Store the states of all \(M\) modes over the training and
      validation windows in \(\mathcal B\) under \(\ell\).
  \ENDIF
\ENDFOR
\FOR{\(k = 0\) \TO \(K - 1\)}
  \STATE Copy from \(\mathcal B[\eta_k]\) the state columns of the modes in
    sector \(k\) into the unit-scale state matrices
    \(\mathbf H_{\mathrm{tr}}^{\mathrm{unit}}\) and \(\mathbf H_{\mathrm v}^{\mathrm{unit}}\).
\ENDFOR
\STATE \(\mathbf H_{\mathrm{tr}} \leftarrow \sigma_* \mathbf H_{\mathrm{tr}}^{\mathrm{unit}}\), \(\mathbf H_{\mathrm v} \leftarrow \sigma_* \mathbf H_{\mathrm v}^{\mathrm{unit}}\).
\STATE \(\boldsymbol\Phi_{\mathrm{tr}} \leftarrow [\tanh(\mathbf H_{\mathrm{tr}} \mathbf Q), \mathbf 1]\),
  \(\boldsymbol\Phi_{\mathrm v} \leftarrow [\tanh(\mathbf H_{\mathrm v} \mathbf Q), \mathbf 1]\)
  by (\ref{eq:features}).
\STATE Obtain \(\mathbf W_{\mathrm{out}}\) from \(\boldsymbol\Phi_{\mathrm{tr}}\)
  and \(\mathbf Y_{\mathrm{tr}}\) with \(\beta\) by (\ref{eq:ridge}).
\RETURN the NRMSE of \(\boldsymbol\Phi_{\mathrm v} \mathbf W_{\mathrm{out}}\)
  against \(\mathbf Y_{\mathrm v}\) by (\ref{eq:nrmse}).
\end{algorithmic}
\end{algorithm}

\paragraph{Associative-Scan State Generation}

The associative scan~\cite{blelloch1990prefix,martin2018parallelizing}
reorganizes the temporal recurrence of the modes at one half-life value so that, with unlimited parallel processors, the number of sequential steps along time grows only logarithmically with the number of updates (a logarithmic span). The state of each
mode is first represented as one complex number
\(\zeta_{j,t} = x^{(1)}_{j,t} + i\, x^{(2)}_{j,t}\), with eigenvalue
\(\lambda_j = r_j e^{i\theta_j}\) and input term
\(d_{j,t} = (w^{(1)}_j + i\, w^{(2)}_j)\, \widetilde u_t\), where \(x^{(1)}_{j,t}\) and \(x^{(2)}_{j,t}\) denote the two coordinates of \(\mathbf x_{j,t}\), and \(w^{(1)}_j\) and \(w^{(2)}_j\) those of \(\mathbf w_j\). The rotation--decay update then becomes the element-wise complex recurrence
\(\zeta_{j,t} = \lambda_j \zeta_{j,t-1} + d_{j,t}\). Each update is an element-wise affine map \(\boldsymbol\zeta \mapsto \mathbf a \odot \boldsymbol\zeta + \mathbf b\) on the vector of the states of all modes, \(\boldsymbol\zeta_t = (\zeta_{1,t}, \ldots, \zeta_{M,t})\), with \(\mathbf a = (\lambda_1, \ldots, \lambda_M)\), the same for every update, and \(\mathbf b = (d_{1,t}, \ldots, d_{M,t})\) at update \(t\), where \(\odot\) denotes the element-wise product, and two such maps compose as
\begin{equation}
(\mathbf a_2, \mathbf b_2) \circ (\mathbf a_1, \mathbf b_1)
= (\mathbf a_2 \odot \mathbf a_1,\;
   \mathbf a_2 \odot \mathbf b_1 + \mathbf b_2),
\label{eq:compose}
\end{equation}
which is associative. An associative scan over these elements therefore
reproduces the serial recurrence, up to floating-point differences. For \(T_{\mathrm{drv}}\) updates (the number of inputs driven in one scan, here the warmup, training, and validation inputs in sequence; line 4 of Algorithm~\ref{alg:candidate-evaluation}), the implemented scan runs in
\(O(\log T_{\mathrm{drv}})\) stages of \(O(T_{\mathrm{drv}} N)\) element-wise
operations, that is, \(O(T_{\mathrm{drv}} N \log T_{\mathrm{drv}})\)
operations in total, with \(O(T_{\mathrm{drv}} N)\) memory for the states.

\paragraph{Across-Candidate State Reuse by Half-Life Value}

The state bank reuses states across SRLR candidates. This reuse is
possible because the SRLR restricts each \(\eta_k\) to the finite set
of half-life values in Table~\ref{tab:search-spaces}. The
\(\gamma\)-normalized design uses the same procedure with its own bank, where line 3 of Algorithm~\ref{alg:candidate-evaluation} uses the flat white-noise PSD. The common-radius search cannot use the bank because its radius takes
continuous values, and the uniform random design is not supported by the bank
in the present implementation. Because the input gains \(g_j\) of
Section~III-A3c are proportional to \(\sigma_*\) and the recurrence is linear,
the states for any \(\sigma_*\) are obtained by computing the states once at
\(\sigma_* = 1\) and multiplying them by \(\sigma_*\). A candidate with
half-life profile \(\boldsymbol\eta\) (Section~III-B2) is then assembled by
selecting, for each sector, the stored state trajectories for the half-life
value assigned to that sector and scaling them by \(\sigma_*\)
(Algorithm~\ref{alg:candidate-evaluation}, lines 8--11). The readout is fitted
and scored as usual, with \(\beta\) changing only the ridge solve. The states
are independent of the model seed, which enters only through the mixing matrix
\(\mathbf Q\) of the feature map (Section~III-A3d). The bank is kept per search so that the searches remain independent across tasks and model seeds, as for the other search configurations (Section~III-B2), and so that the reported cost corresponds to a single task and model seed, as in practical use; sharing a bank across the searches of a system, whose states are identical, would reduce the count further in this benchmark. In each search, the number
of mode trajectories computed thus falls from \(CM\) (all \(M\) modes
recomputed for each of the \(C\) candidates evaluated in that search) to
\(LM\) (computed once for each of the \(L\) distinct half-life values
requested in that search). Here \(L\) is at most the number of allowed
half-life values (six), however large \(C\) is.

\subsubsection{Evaluation Metrics and Aggregation}

For a scored window of length \(T\), with \(\hat y_t\) the prediction and
\(y_t\) the target at time \(t\), the NRMSE is defined as
\begin{equation}
\operatorname{NRMSE}
=
\sqrt{
\frac{\frac{1}{T}\sum_{t=1}^{T} (\hat y_t - y_t)^2}
     {\frac{1}{T}\sum_{t=1}^{T} (y_t - \overline y)^2}},
\qquad
\overline y = \frac{1}{T}\sum_{t=1}^{T} y_t,
\label{eq:nrmse}
\end{equation}
where the mean \(\overline y\) is computed on the same scored window. This is
the root-mean-square error divided by the population standard deviation of the
target on that window. Each case is summarized by the mean and population standard deviation (SD) of the NRMSE across the ten model seeds.

\section{Results}

\subsection{SRLR Versus the Common-Radius Reference}

\begin{figure*}[t]
\centering
\includegraphics[width=\textwidth]{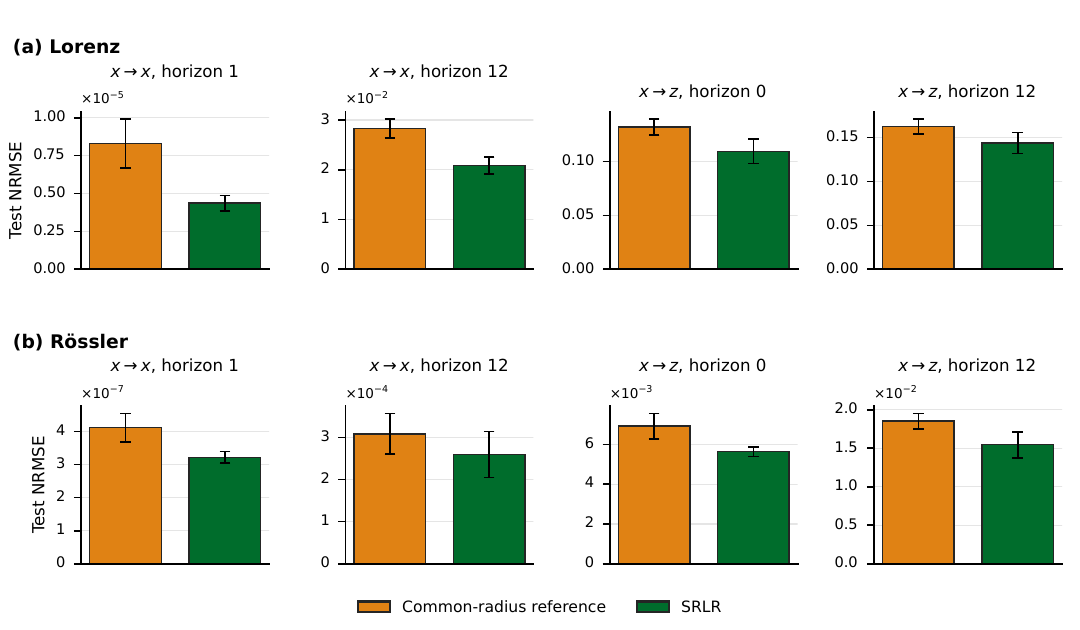}
\caption{Test NRMSE of the SRLR and the common-radius reference on (a)
the four Lorenz tasks and (b) the four R\"ossler tasks. For each system, the
panels show, in order, \(x \to x\) prediction at horizon 1, \(x \to x\)
prediction at horizon 12, \(x \to z\) reconstruction at horizon 0, and
\(x \to z\) reconstruction at horizon 12. Bars show means over ten model
seeds, and error bars show population SD.}
\label{fig:common-sector-ring-results}
\end{figure*}

We first compare the SRLR (\(K = 8\), Section~III-A3b) with its
common-radius reference. Fig.~\ref{fig:common-sector-ring-results} shows the
mean test NRMSE for the eight system--task cases, and
Table~\ref{tab:common_sector_ring_nrmse} reports the corresponding means and
population SDs over the ten model seeds. The SRLR achieved a lower mean NRMSE than the
common-radius reference in all eight cases. The reduction in mean NRMSE relative to the common-radius reference, computed from the unrounded means, ranged from 11.6\% to 47.4\% across the cases.

\begin{table*}[t]
\renewcommand{\arraystretch}{1.15}
\footnotesize
\centering
\caption{Test NRMSE for the common-radius reference and the SRLR: mean
$\pm$ population SD over ten model seeds. The lowest mean in each row is in bold.}
\label{tab:common_sector_ring_nrmse}
\begin{tabular}{lcc}
\hline
Task & Common-radius reference & SRLR \\
\hline
\multicolumn{3}{l}{\textit{Lorenz}} \\
\hline
\rule{0pt}{2.8ex}$x\to x$, horizon 1 & $(8.29 \pm 1.63)\times 10^{-6}$ & $(\mathbf{4.36} \pm 0.507)\times 10^{-6}$ \\
$x\to x$, horizon 12 & $0.0283 \pm 0.00190$ & $\mathbf{0.0209} \pm 0.00167$ \\
$x\to z$, horizon 0 & $0.132 \pm 0.00748$ & $\mathbf{0.109} \pm 0.0112$ \\
$x\to z$, horizon 12 & $0.162 \pm 0.00862$ & $\mathbf{0.144} \pm 0.0121$ \\
\hline
\multicolumn{3}{l}{\textit{R\"ossler}} \\
\hline
\rule{0pt}{2.8ex}$x\to x$, horizon 1 & $(4.12 \pm 0.435)\times 10^{-7}$ & $(\mathbf{3.22} \pm 0.175)\times 10^{-7}$ \\
$x\to x$, horizon 12 & $(3.09 \pm 0.486)\times 10^{-4}$ & $(\mathbf{2.60} \pm 0.549)\times 10^{-4}$ \\
$x\to z$, horizon 0 & $(6.93 \pm 0.645)\times 10^{-3}$ & $(\mathbf{5.65} \pm 0.241)\times 10^{-3}$ \\
$x\to z$, horizon 12 & $0.0185 \pm 0.00101$ & $\mathbf{0.0155} \pm 0.00169$ \\
\hline
\end{tabular}
\end{table*}

\subsection{Comparison With the ESN and LI-ESN Baselines}

\begin{figure*}[t]
\centering
\includegraphics[width=\textwidth]{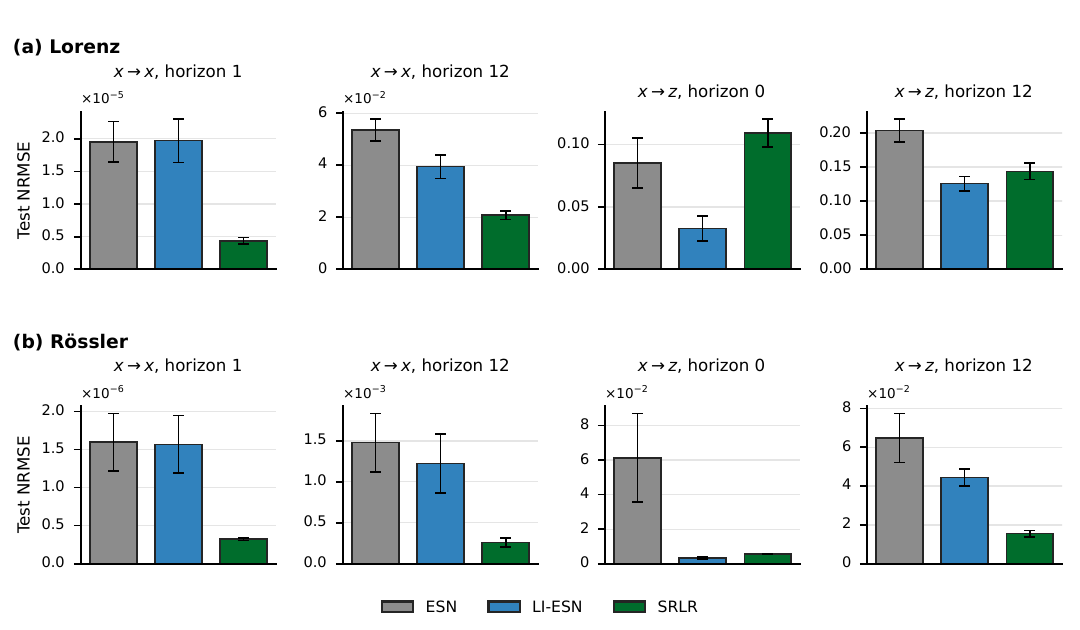}
\caption{Test NRMSE of the ESN, the LI-ESN, and the SRLR on (a) the
four Lorenz tasks and (b) the four R\"ossler tasks, in the task order of
Fig.~\ref{fig:common-sector-ring-results}. Bars show means over ten model
seeds, and error bars show population SD.}
\label{fig:baseline-results}
\end{figure*}

We then compare the SRLR with the ESN and LI-ESN baselines.
Fig.~\ref{fig:baseline-results} shows the mean test NRMSE for the eight
cases, and Table~\ref{tab:esn_li_sector_ring_nrmse} reports the corresponding
means and population SDs over the ten model seeds. The SRLR achieved
the lowest mean NRMSE in five of the eight cases, the LI-ESN in
three, and the ESN in none. The SRLR had the lowest mean in all four \(x \to x\) prediction cases and in the R\"ossler \(x \to z\) reconstruction case at horizon 12, and the LI-ESN had the lowest mean in the two
\(x \to z\) reconstruction cases at horizon 0 and in the Lorenz \(x \to z\)
reconstruction case at horizon 12.

\begin{table*}[t]
\renewcommand{\arraystretch}{1.15}
\footnotesize
\centering
\caption{Test NRMSE for the ESN, the LI-ESN, and the SRLR: mean $\pm$
population SD over ten model seeds. The lowest mean in each row is in bold.}
\label{tab:esn_li_sector_ring_nrmse}
\begin{tabular}{lccc}
\hline
Task & ESN & LI-ESN & SRLR \\
\hline
\multicolumn{4}{l}{\textit{Lorenz}} \\
\hline
\rule{0pt}{2.8ex}$x\to x$, horizon 1 & $(19.5 \pm 3.09)\times 10^{-6}$ & $(19.7 \pm 3.33)\times 10^{-6}$ & $(\mathbf{4.36} \pm 0.507)\times 10^{-6}$ \\
$x\to x$, horizon 12 & $0.0536 \pm 0.00418$ & $0.0395 \pm 0.00456$ & $\mathbf{0.0209} \pm 0.00167$ \\
$x\to z$, horizon 0 & $0.0853 \pm 0.0200$ & $\mathbf{0.0328} \pm 0.0100$ & $0.109 \pm 0.0112$ \\
$x\to z$, horizon 12 & $0.204 \pm 0.0167$ & $\mathbf{0.126} \pm 0.0104$ & $0.144 \pm 0.0121$ \\
\hline
\multicolumn{4}{l}{\textit{R\"ossler}} \\
\hline
\rule{0pt}{2.8ex}$x\to x$, horizon 1 & $(16.0 \pm 3.79)\times 10^{-7}$ & $(15.7 \pm 3.78)\times 10^{-7}$ & $(\mathbf{3.22} \pm 0.175)\times 10^{-7}$ \\
$x\to x$, horizon 12 & $(14.8 \pm 3.58)\times 10^{-4}$ & $(12.3 \pm 3.60)\times 10^{-4}$ & $(\mathbf{2.60} \pm 0.549)\times 10^{-4}$ \\
$x\to z$, horizon 0 & $(61.3 \pm 25.7)\times 10^{-3}$ & $(\mathbf{3.42} \pm 0.636)\times 10^{-3}$ & $(5.65 \pm 0.241)\times 10^{-3}$ \\
$x\to z$, horizon 12 & $0.0648 \pm 0.0127$ & $0.0444 \pm 0.00426$ & $\mathbf{0.0155} \pm 0.00169$ \\
\hline
\end{tabular}
\end{table*}

\subsection{Selected Half-Life Profiles}

\begin{figure*}[p]
\centering
\includegraphics[width=0.96\textwidth]{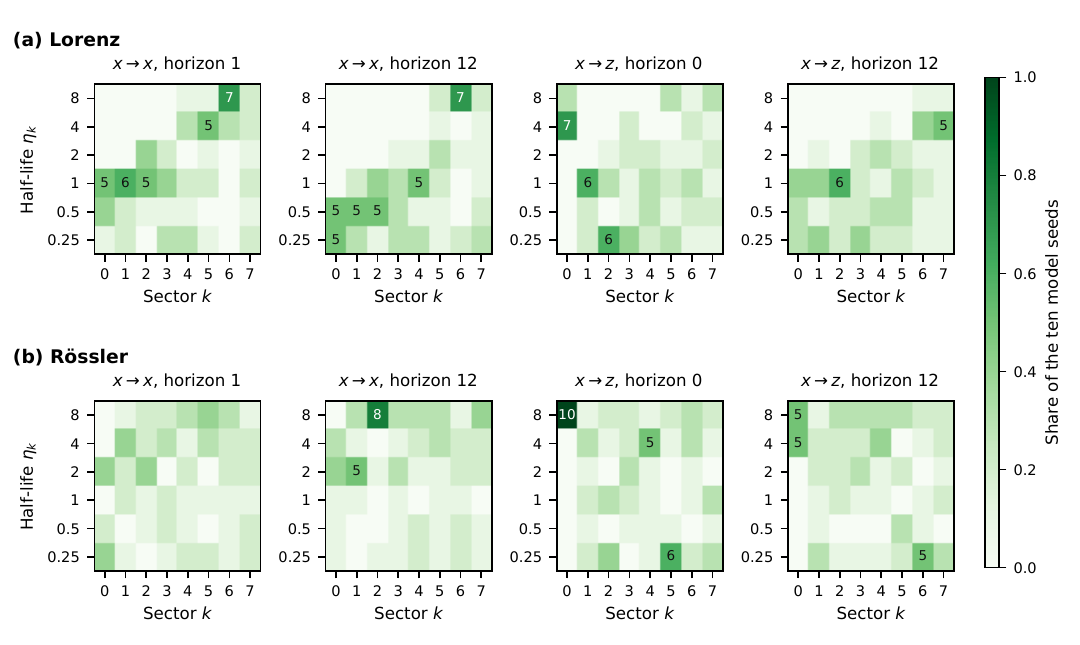}
\caption{Half-life profiles selected by the validation searches for the
SRLR on (a) the four Lorenz tasks and (b) the four R\"ossler tasks, in the task order of Fig.~\ref{fig:common-sector-ring-results}. Each task panel is a
heatmap with the six half-life values
\(\eta_k \in \{0.25, 0.5, 1, 2, 4, 8\}\) (in reservoir updates) as rows and
the sectors 0--7 as columns (sector 0 contains the smallest angles \(\theta_j\), near 0, and sector 7 the largest, near \(\pi\); see (\ref{eq:sector})). The color of a cell indicates the fraction of the ten model seeds that selected that half-life for that sector, and cells selected by at least five seeds are labeled with the seed count.}
\label{fig:sector-ring-half-life-profiles}
\end{figure*}

Fig.~\ref{fig:sector-ring-half-life-profiles} shows the half-life profiles
selected by the validation searches for the SRLR, for the ten model
seeds of each task. Within each system, the tasks share the observed input
sequence, yet the selected profiles changed with the prediction target and
the horizon. The profiles also differed across the ten model seeds.

\subsection{Comparison of Input-Weight Designs}

\begin{figure*}[p]
\centering
\includegraphics[width=0.96\textwidth]{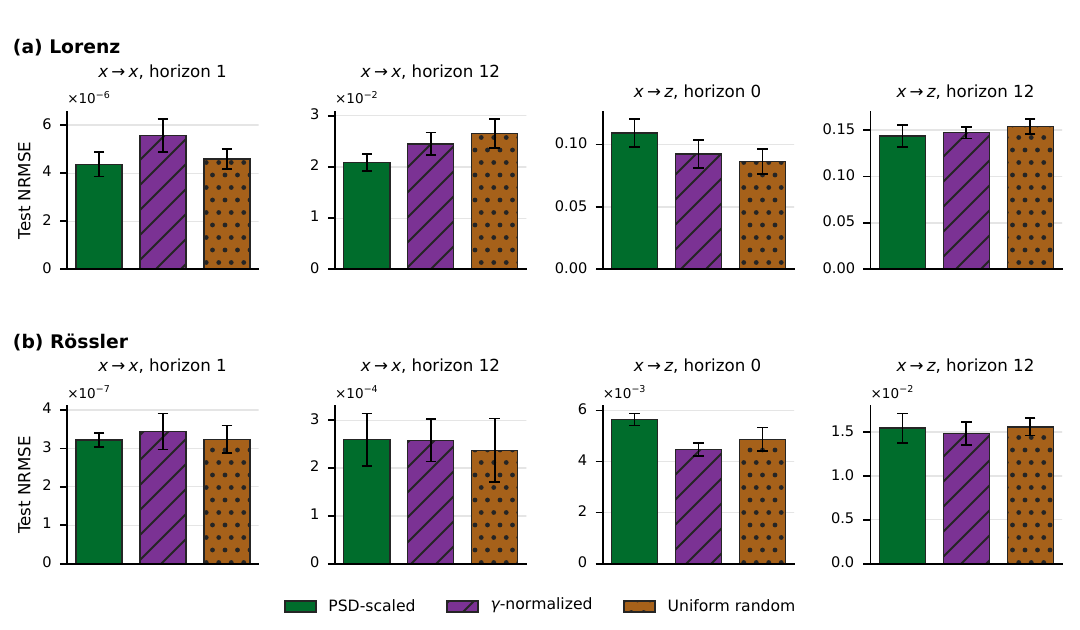}
\caption{Test NRMSE of the three input-weight designs of the SRLR
(PSD-scaled, \(\gamma\)-normalized, and uniform random) on (a) the four Lorenz
tasks and (b) the four R\"ossler tasks, in the task order of
Fig.~\ref{fig:common-sector-ring-results}, each design with the settings
selected by its own search. Bars show means over ten model seeds, and error
bars show population SD.}
\label{fig:input-design-results}
\end{figure*}

We also compare the three input-weight designs of the SRLR
(Section~III-A3c), namely the PSD-scaled, \(\gamma\)-normalized, and uniform
random designs, each with the settings selected by its own search.
Fig.~\ref{fig:input-design-results} shows the mean test NRMSE for the eight
cases, and Table~\ref{tab:input_design_nrmse} reports the corresponding means
and population SDs over the ten model seeds. The PSD-scaled design achieved
the lowest mean NRMSE in four of the eight cases, the
\(\gamma\)-normalized design in two, and the uniform random design in two.
The largest differences among the three designs were in the Lorenz
\(x \to x\) prediction cases at both horizons, where the PSD-scaled design had the lowest mean, and in the \(x \to z\) reconstruction cases of both systems at horizon 0, where it had the highest mean. In these four cases, the highest mean
exceeded the lowest by about 26\% to 28\%. In the remaining four cases, the three means were within 10\% of
one another. The ordering of the three designs thus depended on the task.

\begin{table*}[t]
\renewcommand{\arraystretch}{1.15}
\footnotesize
\centering
\caption{Test NRMSE for the three input-weight designs of the SRLR:
mean $\pm$ population SD over ten model seeds. The lowest mean in each row is in bold.}
\label{tab:input_design_nrmse}
\begin{tabular}{lccc}
\hline
Task & PSD-scaled & \(\gamma\)-normalized & Uniform random \\
\hline
\multicolumn{4}{l}{\textit{Lorenz}} \\
\hline
\rule{0pt}{2.8ex}$x\to x$, horizon 1 & $(\mathbf{4.36} \pm 0.507)\times 10^{-6}$ & $(5.56 \pm 0.691)\times 10^{-6}$ & $(4.59 \pm 0.424)\times 10^{-6}$ \\
$x\to x$, horizon 12 & $\mathbf{0.0209} \pm 0.00167$ & $0.0245 \pm 0.00222$ & $0.0265 \pm 0.00286$ \\
$x\to z$, horizon 0 & $0.109 \pm 0.0112$ & $0.0923 \pm 0.0113$ & $\mathbf{0.0864} \pm 0.00990$ \\
$x\to z$, horizon 12 & $\mathbf{0.144} \pm 0.0121$ & $0.147 \pm 0.00614$ & $0.154 \pm 0.00806$ \\
\hline
\multicolumn{4}{l}{\textit{R\"ossler}} \\
\hline
\rule{0pt}{2.8ex}$x\to x$, horizon 1 & $(\mathbf{3.22} \pm 0.175)\times 10^{-7}$ & $(3.44 \pm 0.469)\times 10^{-7}$ & $(3.23 \pm 0.360)\times 10^{-7}$ \\
$x\to x$, horizon 12 & $(2.60 \pm 0.549)\times 10^{-4}$ & $(2.58 \pm 0.441)\times 10^{-4}$ & $(\mathbf{2.37} \pm 0.666)\times 10^{-4}$ \\
$x\to z$, horizon 0 & $(5.65 \pm 0.241)\times 10^{-3}$ & $(\mathbf{4.48} \pm 0.247)\times 10^{-3}$ & $(4.88 \pm 0.464)\times 10^{-3}$ \\
$x\to z$, horizon 12 & $0.0155 \pm 0.00169$ & $\mathbf{0.0148} \pm 0.00131$ & $0.0156 \pm 0.000987$ \\
\hline
\end{tabular}
\end{table*}

\subsection{Scale of the States of the Modes Under the Three Input-Weight Designs}

Next, we examine how the three input-weight designs affect the scale of the
state of each mode, using the profiles selected for the PSD-scaled design in
Section~IV-C as fixed settings. For each of the 40 task-by-seed profiles per
system, the recurrence and the observed input sequence are fixed, only the
input weights change, and the states of the modes are measured before the
orthogonal mixing, the feature map, the readout, and the scoring. Each
run starts from the zero state, drives the reservoir with the 2,000
warmup inputs without scoring, and retains the 10,000 states of the training
window. For each mode \(j = 1, \ldots, 32\), the two state coordinates are
treated separately, and the scale of each coordinate is its population SD
over the retained states,
\(s_{j,c} = \sqrt{\operatorname{Var}(x^{(c)}_{j,t})}\), \(c \in \{1, 2\}\).
This gives 64 coordinate scales per profile under each design. Each profile
under each design is summarized by the ratio of the 90th to the 10th
percentile (P90/P10) of its 64 coordinate scales, with both percentiles
computed by linear interpolation. Because a gain common to all modes (\(\sigma_*\) for the PSD-scaled and \(\gamma\)-normalized designs and \(s_{\mathrm{in}}\) for the uniform random design) multiplies all 64 coordinate scales alike, the ratio does not depend on its value. A ratio close to 1 indicates little
relative variation in coordinate scale between P10 and P90 within the
profile.

\begin{figure}[!tp]
\centering
\includegraphics[width=\columnwidth]{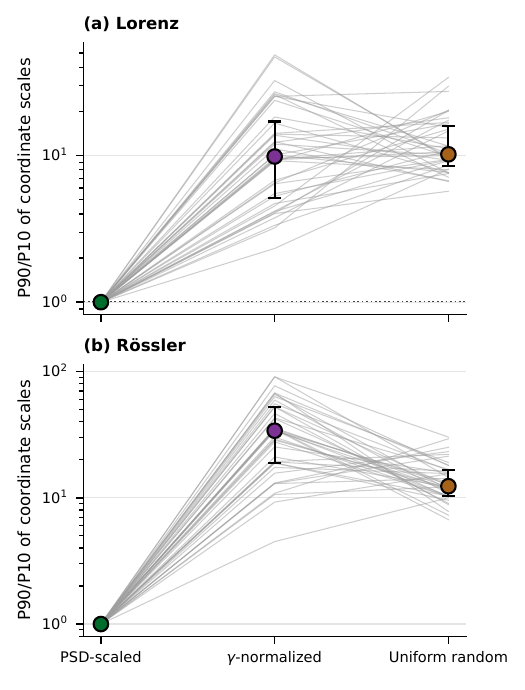}
\caption{Ratio of the 90th to the 10th percentile (P90/P10) of the 64 coordinate scales within each profile, where each scale is the population SD of one coordinate of the states of the modes, for
the PSD-scaled, \(\gamma\)-normalized, and uniform random input-weight designs
(Section~III-A3c) on (a) the Lorenz tasks and (b) the R\"ossler tasks. Each thin line connects the values of one profile (40 profiles, one per task and model seed) across the three designs.
Large outlined symbols mark medians, and vertical bars span the interquartile
range. The dotted line marks a ratio of 1.}
\label{fig:input-design-mode-scale}
\end{figure}

Fig.~\ref{fig:input-design-mode-scale} shows this ratio. The PSD-scaled
design gave a lower ratio than the other two designs in all 40 Lorenz
profiles and all 40 R\"ossler profiles. The median ratio was close
to 1 for the PSD-scaled design (1.00 for both systems), compared with 9.86 and 10.2 on Lorenz and 34.0 and 12.3 on R\"ossler for the \(\gamma\)-normalized and uniform random designs, respectively.

\begin{figure*}[!tp]
\centering
\includegraphics[width=\textwidth]{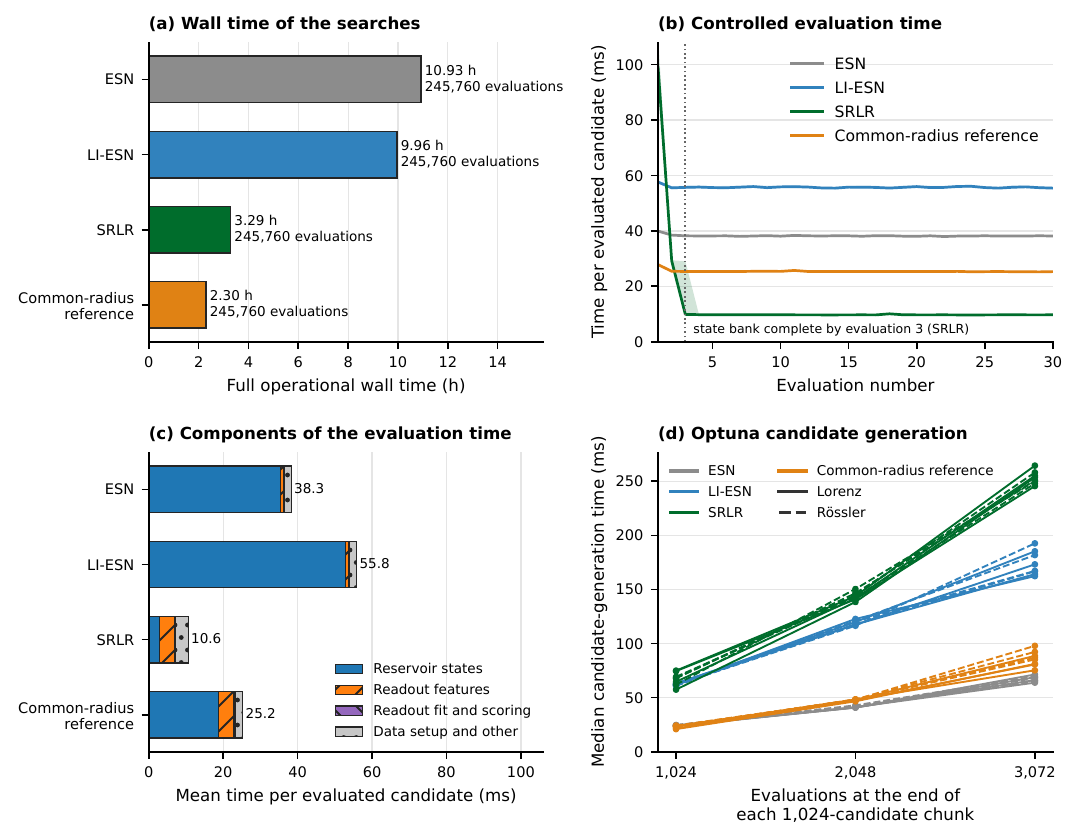}
\caption{Computational cost of the validation searches. (a) Full operational
wall time and number of evaluations that returned a finite validation NRMSE
for the SRLR, the ESN, the LI-ESN, and the common-radius reference (80
searches per search configuration). The searches of the four configurations
ran sequentially, the ten seeds of each case ran concurrently, and other
workloads on the machine were not controlled. (b) Time per evaluated candidate for the first 30 of the 128 replayed candidates (the first 128 candidates recorded in each search, re-evaluated in their recorded order) in each of 80 searches per search configuration, one per
system, task, and model seed, measured with one worker process and one BLAS
thread on one machine, using a fresh process for each search and, for the
SRLR, a fresh state bank per search. Lines show medians across
searches, and shading covers the middle 50\% (25th--75th percentile range).
For the SRLR, all six half-life values of the state bank had been computed by the third evaluation in all 80 searches, and a half-life value
computed in an earlier evaluation could be reused before the bank was
complete. (c) Mean evaluation-time components from the same
measurements as panel (b), over all 128 replayed candidates (10,240 per search configuration). (d) Median time for Optuna to generate the next candidate before evaluation, recorded during the searches, over successive blocks of 1,024 candidates, separately for each case and for
the four search configurations of panel (a) (10,240 measurements per point,
1,024 candidates \(\times\) ten seeds). Solid and dashed lines denote Lorenz and R\"ossler, respectively.}
\label{fig:computational-cost}

% Keep the timing-scope table with its figure, retaining the native table caption.
\par\medskip
\begingroup
\makeatletter
\def\@captype{table}
\makeatother
\footnotesize
\centering
\caption{Operations included in the timing measurements reported in
Fig.~\ref{fig:computational-cost}. ``Measured'' denotes the interval shown
directly, whereas ``Included'' denotes work contained in the full operational wall time of panel~(a), excluding time spent writing summaries after the
recorded end of the searches.}
\label{tab:computational-cost-scopes}
\renewcommand{\arraystretch}{1.15}
\begin{tabular}{@{}p{0.29\textwidth}p{0.20\textwidth}p{0.23\textwidth}p{0.20\textwidth}@{}}
\hline
Operation group & (a) Full operational wall time &
(b)--(c) Evaluation timing & (d) Candidate generation \\
\hline
Optuna candidate generation & Included & --- & Measured \\
Candidate evaluation & Included & Measured in (b); decomposed in (c) & --- \\
Result reporting and storage & Included & --- & --- \\
\raggedright Worker and process management, checkpoints, and finalization of the searches &
Included & --- & --- \\
\hline
\end{tabular}
\endgroup
\end{figure*}

\subsection{Computational Cost of the Candidate Search}

Finally, we report the computational cost of the validation searches
(Fig.~\ref{fig:computational-cost}). Each of the four search
configurations in Fig.~\ref{fig:computational-cost}(a), namely the
SRLR, the ESN, the LI-ESN, and the common-radius reference, evaluated
245,760 candidates in total (80 searches of 3,072 evaluations each), and every
evaluation returned a finite validation NRMSE. The searches ran on a
14-core Apple M4 Pro processor, with the ten model seeds of each case run
concurrently. Their full operational wall time was 3.29~h for the
SRLR, 10.93~h for the ESN, 9.96~h for the LI-ESN, and 2.30~h for the
common-radius reference. Over the 80 SRLR searches,
recomputing all 32 modes for every candidate would have required
\(M \sum C = \) 7,864,320 mode trajectories, where the sum runs over
the 80 searches. The state banks computed \(M \sum L = \) 15,360, that
is, 32 for each of the \(L \le 6\) half-life values requested in each search
(\(L = 6\) in every search), a reduction of 99.8\%
(Section~III-B3b).

In the separate controlled measurements of
Fig.~\ref{fig:computational-cost}(b) and (c), made on the same machine with
one worker process and one Basic Linear Algebra Subprograms (BLAS) thread,
one evaluated candidate took on average 10.6~ms for the SRLR,
38.3~ms for the ESN, 55.8~ms for the LI-ESN, and 25.2~ms for the
common-radius reference. The largest
difference among the components was in the reservoir-state stage,
which took 2.8~ms for the SRLR, 35.3~ms for the ESN, and 52.9~ms for the LI-ESN. The construction of the readout features, the largest component of the SRLR evaluation, took 4.1~ms. In Fig.~\ref{fig:computational-cost}(d), the
time for Optuna to generate the next candidate increased as the searches
accumulated candidates, most strongly for the SRLR. The full operational wall time of panel (a) also includes this candidate-generation time (Table~\ref{tab:computational-cost-scopes}), which was longer for the SRLR than for the common-radius reference in every block; this contributes to the shorter full wall time of the common-radius reference than of the SRLR despite its longer time per evaluated candidate.
Table~\ref{tab:computational-cost-scopes} summarizes the operations timed in
panels (a), (b)--(c), and (d).

\newpage
\section{Discussion}

\subsection{Benchmark Outcomes and Task-Dependent Timescales}

The two comparisons show that allocating the decay timescale sector by sector is a useful degree of freedom, and that the SRLR with this freedom is, as a whole, competitive with the nonlinear baselines. First, the SRLR achieved a lower mean test NRMSE than the common-radius reference in all eight system--task cases (Fig.~\ref{fig:common-sector-ring-results}). Because the two differ only in whether the decay timescale may vary across sectors, this difference can be attributed to that freedom. Second, in the comparison with the ESN and LI-ESN, the lowest mean NRMSE was achieved by the SRLR in five of the eight cases, by the LI-ESN in three, and by the ESN in none (Fig.~\ref{fig:baseline-results}). The SRLR, whose recurrence is linear, is therefore competitive with the separately tuned nonlinear baselines.

Suitable reservoir dynamics are task-dependent, even for a single input sequence. Within each system, the four tasks share the observed input sequence, and the validation-selected half-life profiles changed with the prediction target and the horizon (Fig.~\ref{fig:sector-ring-half-life-profiles}). The two nonlinear baselines show the same dependence. Leaky integration, which distinguishes the LI-ESN from the ESN, is a standard way to adjust the timescale of the state update of an ESN to the task~\cite{jaeger2007leaky} and can substantially improve prediction in some tasks~\cite{ebato2024impact}. The mean-error gap between the separately tuned ESN and LI-ESN, however, differed across the target--horizon cases, and in the two \(x \to x\) prediction cases at horizon 1 their means were close (Fig.~\ref{fig:baseline-results}, Table~\ref{tab:esn_li_sector_ring_nrmse}). Both patterns are consistent with the prior observation that, under a shared input, different prediction targets require and select different reservoir timescales~\cite{tanaka2022diversetimescales}, and with the view that the useful fading memory (the memory of past inputs, which decays with time) is set by the target computation~\cite{carroll2022optimizing}.

\subsection{Input-Weight Scaling, State Scale, and Prediction Performance}

The PSD scaling of the input weights was introduced to equalize the scale of the states of the modes across candidates, so that the scale of the states of the modes does not change with the eigenvalue distribution among candidates. In Section~IV-E, we tested how closely the realized states follow this design by fixing the recurrence, the input sequence, and the model seed and changing only the input weights. In all 40 Lorenz and all 40 R\"ossler profiles, the PSD-scaled design gave a lower P90/P10 ratio than the other two designs, with median ratios close to 1 (Fig.~\ref{fig:input-design-mode-scale}). The PSD scaling therefore suppresses the variation in coordinate scale, so that differences in scale do not enter the comparison of decay-timescale profiles.

In terms of prediction performance, the comparison in Section~IV-D shows how the three designs differ. The PSD-scaled design achieved the lowest mean NRMSE in four of the eight cases, and among these four the differences among the three designs were largest in the Lorenz \(x \to x\) prediction cases. In the \(x \to z\) reconstruction cases at horizon 0, by contrast, it had the highest mean, about 26\% above the lowest of the three means (Fig.~\ref{fig:input-design-results}). This indicates that equalizing the scale of the states of the modes does not by itself determine accuracy: the three designs distribute the input differently across the modes, and the distribution that suited each task differed. The role of the PSD scaling that the present results support is therefore the one tested in Section~IV-E, namely keeping differences in scale out of the comparison of decay-timescale profiles. Future research directions include allocating the input gain of each mode according to the prediction target or the horizon, as in task-tailored reservoir design~\cite{aceituno2020tailoring}.

\subsection{Seed-Level Variability and Unresolved Contributors}

In addition to varying with the prediction target and the horizon, the selected profiles differed across the ten model seeds (Fig.~\ref{fig:sector-ring-half-life-profiles}). Because each model seed generates its own fixed orthogonal mixing matrix \(\mathbf Q\) and runs its own finite TPE search, both can contribute to this variation. The present evidence does not separate the two contributions, and how each of them affects the selected profiles and the prediction performance is a question for future research.

\subsection{Computational Cost and Scalability}

Because the search for the SRLR has ten coordinates, eight of them the sector half-lives, against three for the ESN and four for the LI-ESN, evaluating candidates at low computational cost was a practical concern. In the recorded searches, the SRLR, the ESN, and the LI-ESN each evaluated 245,760 candidates, and the SRLR did so in 3.29~h, compared with 10.93~h for the ESN and 9.96~h for the LI-ESN (Fig.~\ref{fig:computational-cost}(a)). In the controlled per-candidate measurement, the SRLR also took 10.6~ms per candidate, against 38.3~ms for the ESN and 55.8~ms for the LI-ESN (Fig.~\ref{fig:computational-cost}(b) and (c)). The difference is concentrated in the reservoir-state stage (Fig.~\ref{fig:computational-cost}(c)), which is consistent with the implemented candidate evaluation (Algorithm~\ref{alg:candidate-evaluation}). It computes the linear recurrence by an associative scan~\cite{blelloch1990prefix,martin2018parallelizing}, as in recent efficient linear-recurrence models~\cite{orvieto2023resurrecting,pinna2026paralesn}, and reuses the states for each half-life value across candidates, so that the recurrence is not recomputed for every candidate. Over the 80 searches, the state banks reduced the number of computed mode trajectories by 99.8\% relative to per-candidate recomputation (Section~IV-F).

The time for Optuna to generate the next candidate also increased as the searches progressed, most strongly for the SRLR (Fig.~\ref{fig:computational-cost}(d)). Because TPE builds its proposal distribution for each search coordinate from the accumulated evaluation history~\cite{bergstra2011algorithms}, both the length of that history and the number of search coordinates contribute to the cost of generating a candidate. The observed growth is consistent with this mechanism. For searches over many candidates, efficient candidate generation is therefore as important as efficient candidate evaluation.

The present evaluation used a reservoir of \(N = 64\) units, and extending it to larger \(N\) will require further cost reduction. With the state computation reduced, the next bottleneck is the construction of the readout features (the largest component of the SRLR evaluation, 4.1~ms, Fig.~\ref{fig:computational-cost}(c)), in which applying the dense \(N \times N\) orthogonal matrix \(\mathbf Q\) at every time step costs \(O(TN^2)\). One option is to use Hadamard matrices. A composition of Hadamard and diagonal matrices, applied via the fast Hadamard transform, reduces this cost to \(O(TN \log N)\)~\cite{yu2016orthogonal}. Whether such a replacement preserves the accuracy of the SRLR is an open problem.

\subsection{Scope and Future Directions}

These conclusions are based on eight system--task cases, two dynamical systems with four tasks each, evaluated over ten model seeds under a single protocol, with the SRLR at \(K = 8\). Future research directions include evaluating the SRLR on a wide range of systems and on further prediction targets and horizons defined over a shared input. They also include relating the decay timescales selected by the SRLR to the timescales selected by nonlinear reservoirs such as the LI-ESN.

To make the eigenvalue distribution the design variable, the SRLR restricts its recurrence to one block-diagonal construction that is normal (it commutes with its transpose), whereas the randomly generated recurrences of standard ESNs are typically non-normal. Non-normal structure can introduce transient amplification through non-orthogonal modes~\cite{hennequin2012nonnormal,kerg2019nonnormal}, a behavior that the present parameterization does not cover. How sector-wise decay-timescale allocation can be extended to non-normal recurrences is an open problem.

\section{Conclusion}

In this paper, we proposed the Sector-Ring Linear Reservoir (SRLR), which makes the eigenvalue distribution of its linear recurrence a design variable by dividing the eigenvalue angles into sectors and assigning one decay timescale, expressed as a half-life, to each sector. To compare candidate eigenvalue distributions with the scale of each mode's state kept equal, the input weights are scaled using the power spectrum of the training input. To evaluate candidates at low computational cost, the states for each half-life value are reused across candidates. In two comparisons on prediction and reconstruction tasks built from chaotic systems, the SRLR achieved a lower mean test NRMSE than its common-radius reference in all eight cases, which supports sector-wise allocation of the decay timescale as a useful degree of freedom. It was also competitive with the nonlinear ESN and LI-ESN baselines. The selected half-life profiles changed with the prediction target and the horizon, showing that suitable reservoir dynamics are task-dependent. The PSD scaling of the input weights kept the scale of the states of the modes close to equal within each selected profile, so that differences in scale did not enter the comparison of decay-timescale profiles, and its effect on accuracy depended on the task. Compared with two alternative input-weight designs, the PSD-scaled design achieved the lowest mean NRMSE in four of the eight cases. The state reuse reduced the number of computed mode trajectories in the search by 99.8\% relative to per-candidate recomputation. Future research will address a wider range of systems and tasks, the extension to non-normal recurrences, and further cost reduction for larger reservoirs.

\appendices

\section{Input-Weight Scaling}
\label{app:input-weight-scaling}

This appendix derives the input-weight scaling used in Section~III-A3c:
for each block, the direction \(\phi_j\) that equalizes the predicted variances
of the two coordinates of the mode's state, and the gain \(g_j\) that sets them to
the target variance \(\sigma_*^2\). The scaling uses the training
input only through its one-sided PSD. The target scale \(\sigma_*\) is one of the three settings of an SRLR candidate and is selected on the validation window (Section~III-B2).

Because the block recurrence of Section~III-A3a acts on the input as a linear
filter, its predicted response can be written in the frequency domain~\cite{oppenheim1999discrete}. Let
\[
\begin{aligned}
z(f) &= \exp\!\left(-i \frac{2\pi f}{f_s}\right), \\
\mathbf G_j(f) &= \left[\mathbf I - z(f)\mathbf R_j\right]^{-1},
& 0 \le f \le \frac{f_s}{2}.
\end{aligned}
\]
For unit gain and the direction
\(\mathbf q(\phi) = [\cos\phi, \sin\phi]^\mathsf T\), the predicted state
covariance of block \(j\) is
\[
\mathbf C_j(\phi) = \operatorname{Re} \int_0^{f_s/2} \widehat P_u(f)\,
\mathbf G_j(f) \mathbf q(\phi) \mathbf q(\phi)^\mathsf{T}
\mathbf G_j(f)^* \, df,
\]
where \(^*\) denotes the conjugate transpose. Here \(\widehat P_u\) is a
nonnegative one-sided Welch PSD~\cite{welch1967use} of the training input
after standardization by the training-window mean and SD,
computed with a Hann window, 1,024-sample segments, 50\% overlap, and constant
segment-mean detrending, with no further centering applied to the full series.
The sampling rate is normalized to \(f_s = 2\pi\), and the density is rescaled
so that its trapezoidal integral equals the population variance of the same
standardized input.

Writing
\[
\begin{aligned}
a(f) &= 1 - z(f) r_j \cos\theta_j, \\
b(f) &= z(f) r_j \sin\theta_j, \\
D(f) &= a(f)^2 + b(f)^2,
\end{aligned}
\]
the filtered direction is
\[
\mathbf G_j(f) \mathbf q(\phi) = \frac{1}{D(f)}
\begin{bmatrix}
a(f)\cos\phi - b(f)\sin\phi \\
b(f)\cos\phi + a(f)\sin\phi
\end{bmatrix},
\]
and with the three integrals
\[
\begin{aligned}
S_j &= \int_0^{f_s/2} \widehat P_u(f)
       \frac{|a|^2 + |b|^2}{|D|^2} df, \\
\Delta_j &= \int_0^{f_s/2} \widehat P_u(f)
       \frac{|a|^2 - |b|^2}{|D|^2} df, \\
X_j &= \int_0^{f_s/2} \widehat P_u(f)
       \frac{\operatorname{Re}\{a \overline b\}}{|D|^2} df,
\end{aligned}
\]
in which the integrand arguments \((f)\) are omitted for brevity, the two unit-gain
coordinate variances reduce to
\[
\begin{aligned}
V_{1,j}(\phi) &= \frac{S_j}{2} + \frac{\Delta_j}{2}\cos 2\phi
                 - X_j \sin 2\phi, \\
V_{2,j}(\phi) &= \frac{S_j}{2} - \frac{\Delta_j}{2}\cos 2\phi
                 + X_j \sin 2\phi.
\end{aligned}
\]

The variances are equal whenever the \(2\phi\)-sinusoid vanishes, and a
deterministic choice is
\[
\phi_j = \frac{1}{2}\left(
\operatorname{atan2}(-\Delta_j, -2X_j) \bmod \pi\right).
\]
For this direction,
\(V_{1,j}(\phi_j) = V_{2,j}(\phi_j) = S_j/2\); the unit-gain predicted
variance of Section~III-A3c is therefore \(\kappa_j = S_j/2\), and the gain
\[
g_j = \sigma_* \sqrt{\frac{2}{S_j}}
= \frac{\sigma_*}{\sqrt{\kappa_j}}
\]
gives both predicted coordinate variances the target value
\(\sigma_*^2\).

The derivation requires \(0 < r_j < 1\), an integrable nonnegative
\(\widehat P_u\), \(D(f) \ne 0\) on the integration interval, and
\(S_j > 0\), with \(\sigma_* > 0\). If \(\Delta_j = X_j = 0\), every direction
equalizes the variances. The scaling concerns only the
predicted variances of the state coordinates before the mixing by
\(\mathbf Q\), and does not imply whitening of the realized features.

\section{Evaluation Counts and Validation Search Progress}
\label{app:validation-search-progress}

% Keep the taller table and both appendix figures on their shared float page.
\setlength{\dblfloatsep}{3pt}
\setlength{\dbltextfloatsep}{6pt}

\begin{table*}[!t]
\renewcommand{\arraystretch}{1.15}
\footnotesize
\centering
\caption{Final evaluation counts of the validation searches for each search
configuration and system--task case. Each entry is the count shared by the ten
model seeds of that case. All 48 configuration--case pairs met the stopping
rule of Section~III-B2 before reaching the budget of 8,192 evaluations.}
\label{tab:final-evaluation-counts}
\begin{tabular}{lrrrrrr}
\hline
Task & ESN & LI-ESN & SRLR (PSD-scaled) & Common-radius reference & SRLR (\(\gamma\)-normalized) & SRLR (uniform random) \\
\hline
\multicolumn{7}{l}{\textit{Lorenz}} \\
\hline
$x\to x$, horizon 1 & 3,072 & 3,072 & 3,072 & 3,072 & 3,072 & 4,096 \\
$x\to x$, horizon 12 & 3,072 & 3,072 & 3,072 & 3,072 & 3,072 & 3,072 \\
$x\to z$, horizon 0 & 3,072 & 3,072 & 3,072 & 3,072 & 3,072 & 3,072 \\
$x\to z$, horizon 12 & 3,072 & 3,072 & 3,072 & 3,072 & 3,072 & 3,072 \\
\hline
\multicolumn{7}{l}{\textit{R\"ossler}} \\
\hline
$x\to x$, horizon 1 & 3,072 & 3,072 & 3,072 & 3,072 & 3,072 & 4,096 \\
$x\to x$, horizon 12 & 3,072 & 3,072 & 3,072 & 3,072 & 3,072 & 4,096 \\
$x\to z$, horizon 0 & 3,072 & 3,072 & 3,072 & 3,072 & 3,072 & 3,072 \\
$x\to z$, horizon 12 & 3,072 & 3,072 & 3,072 & 3,072 & 4,096 & 3,072 \\
\hline
\end{tabular}
\end{table*}

\begin{figure*}[!t]
\centering
\includegraphics[width=\textwidth]{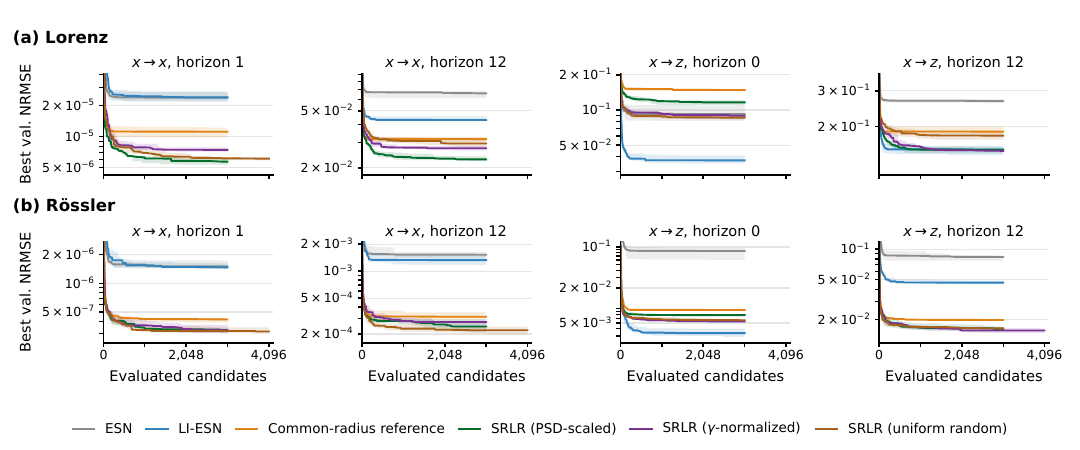}
\caption{Best validation NRMSE found so far versus the number of evaluated
candidates for the six search configurations on the eight system--task cases
((a) Lorenz, (b) R\"ossler). Lines show the median across
the ten model seeds, and bands span the interquartile range. Each curve ends at
the evaluation count in Table~\ref{tab:final-evaluation-counts}. The vertical axis is truncated at the top so that the later part of the searches is visible, and the first evaluations of each curve lie above the plotted range.}
\label{fig:validation-search-progress}
\end{figure*}

\begin{figure*}[!t]
\centering
\includegraphics[width=\textwidth]{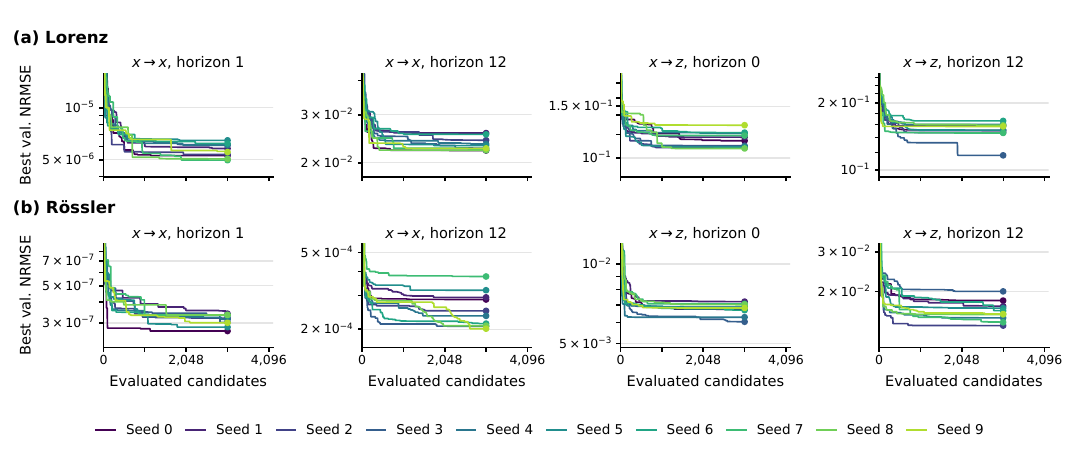}
\caption{Best validation NRMSE found so far versus the number of evaluated
candidates for the SRLR with the PSD-scaled design, shown separately
for the ten model seeds ((a) Lorenz, (b) R\"ossler). Each
line ends at the evaluation count of its case in
Table~\ref{tab:final-evaluation-counts}, and a marker shows the final value of
each line. The vertical axis is truncated at the top as in
Fig.~\ref{fig:validation-search-progress}.}
\label{fig:validation-search-progress-seeds}
\end{figure*}

This appendix reports the evaluation count at which the searches of each
search configuration and system--task case stopped under the stopping rule of
Section~III-B2 (Table~\ref{tab:final-evaluation-counts}), and the optimization
histories of the searches (Figs.~\ref{fig:validation-search-progress} and
\ref{fig:validation-search-progress-seeds}). Because the ten model seeds of a
case stop at the same count, each entry of
Table~\ref{tab:final-evaluation-counts} is a single count. All 48 configuration--case pairs met the stopping rule before reaching the budget of 8,192 evaluations. Forty-four pairs stopped at 3,072 evaluations, the first
count at which the rule is checked, and four at 4,096. The four were the
uniform random design on Lorenz \(x \to x\) at horizon 1 and on R\"ossler
\(x \to x\) at horizons 1 and 12, and the \(\gamma\)-normalized design on
R\"ossler \(x \to z\) at horizon 12.

Fig.~\ref{fig:validation-search-progress} shows, for each case, the median over
the ten model seeds of the best validation NRMSE found so far, \(E^*(n)\) of
Section~III-B2, as a function of the number \(n\) of evaluated candidates,
with the interquartile range shown as a band, for the six search configurations. The median best validation NRMSE decreased
early in the searches, and later improvements were smaller. The stopping
statistic of Section~III-B2 is the median over seeds of the per-seed relative
improvement \(\delta(n)\), which differs from the relative improvement of the
median curve in Fig.~\ref{fig:validation-search-progress}. For example, for
the SRLR with the PSD-scaled design on R\"ossler \(x \to x\) at
horizon 1, the median curve decreased by 3.61\% between 2,048 and 3,072
evaluations, whereas the median of the per-seed improvements was 0.488\%,
below the threshold of 0.5\%.

Fig.~\ref{fig:validation-search-progress-seeds} shows the same progression for
the SRLR with the PSD-scaled design, separately for each model seed.
Within a case, the ratio of the highest to the lowest final best validation
NRMSE across the ten seeds ranged from 1.16 to 1.86 across the eight cases, and the rate of improvement also varied across seeds. The selected profiles likewise
differed across seeds (Section~IV-C).

\clearpage
\section*{Acknowledgment}
The authors used Claude Fable 5.1 (Anthropic) through Claude Code and GPT-5.6 Sol and GPT-6 Astra (OpenAI) through Codex to assist in drafting and editing the text, in typesetting, and in writing the experiment and figure-plotting code. All content was reviewed and verified by the authors, who take full responsibility for it.

\bibliographystyle{IEEEtran}
\bibliography{references}

\EOD

\end{document}